\documentclass[11pt]{article}

\usepackage[preprint]{acl}

\usepackage{times}
\usepackage{latexsym}

\usepackage[T1]{fontenc}

\usepackage[utf8]{inputenc}

\usepackage{microtype}

\usepackage{inconsolata}

\usepackage{graphicx}
\usepackage{xcolor}
\usepackage{booktabs}
\usepackage{amssymb}
\usepackage{amsmath}

\newcommand{\nk}[1]{\textcolor{black}{#1}}

\newcommand{\sd}[1]{\,{\scriptsize(#1)}}

\title{When to Ask, When to Answer: Strategic Clarification for Ambiguous Reference Resolution in Dialogue}

\author{Ibrahim Khebour \\
  Colorado State University \\
  \texttt{ibrahim.khebour@colostate.edu} \\\And
  Nikhil Krishnaswamy \\
  Colorado State University \\
  \texttt{nikhil.krishnaswamy@colostate.edu} \\}

\begin{document}
\maketitle
\begin{abstract}
For humans, referential disambiguation is a collaborative process involving clarification questions asked when they will meaningfully change the outcome. However, large language models (LLMs) have been shown to be biased toward premature answering, preferring definite single-turn responses over more clarification exchanges. In this paper, we study the problem of ``strategic clarification''---asking clarification questions in a reference game when their answers will narrow the set of candidate referents. We construct a controlled test environment from the SIMMC~2.1 dataset that records how many questions each ambiguous reference needs to be successfully resolved, and examine how baseline LLM behavior and common post-training methods ask clarifying questions vs. attempt to answer the referent. We also propose {\bf BACE (Belief-Aware Critic with Epistemic tracking)}, a lightweight action chooser method on top of a frozen LM backbone, and show that BACE more consistently chooses the optimal number of clarifying questions than methods such as GRPO and ArCHer.

\end{abstract}

\vspace*{-2mm}
\section{Introduction}
\vspace*{-2mm}

\begin{figure}[!t]
  \centering
  \includegraphics[width=0.85\columnwidth]{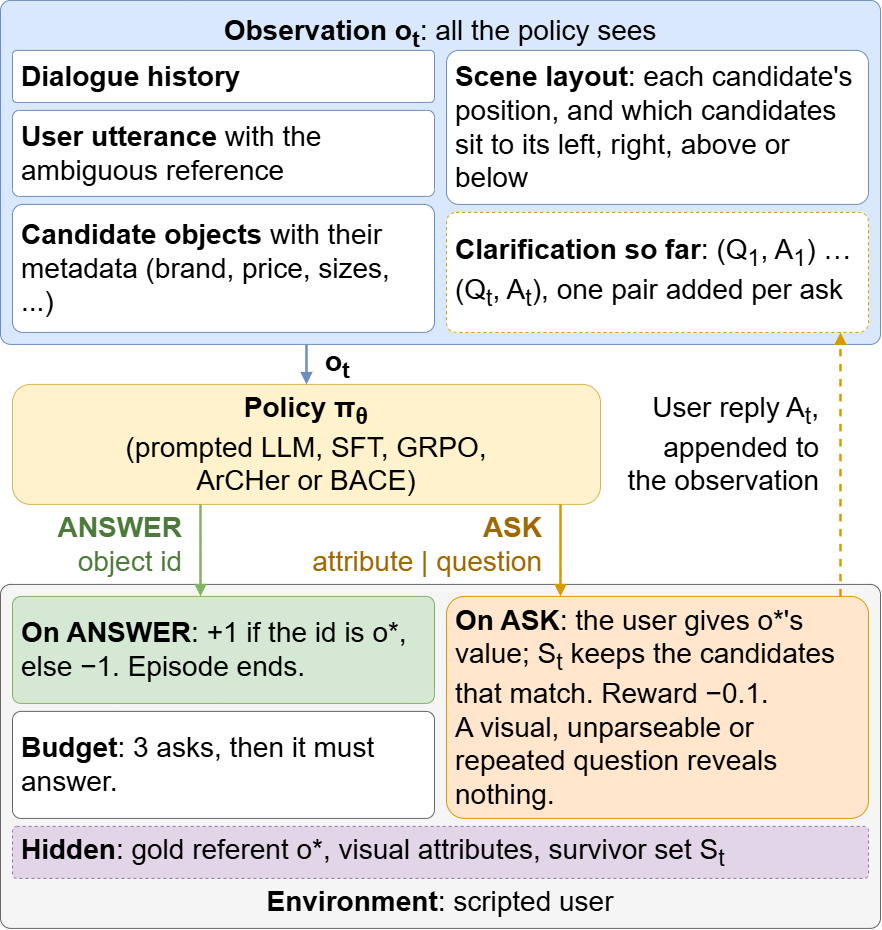}
\vspace*{-2mm}
  \caption{Strategic Clarification. At each turn the policy either asks about one attribute, at a cost of 0.1, or answers with an object ID, and the scripted user replies with the gold referent's value.}
  \label{fig:task}
\vspace*{-2mm}
\end{figure}

\nk{If a person enters a room full of jackets and asks ``What size is that jacket?'', the natural response is to ask which jacket they mean. Previous studies have shown how current large language models (LLMs), which were trained to answer the original prompt quickly and proactively, tend to choose one of the candidate jackets as the golden referent, without asking the logical question.}

People understand that ambiguous reference resolution is a collaborative process
\citep{clark1986referring, clark1991grounding}. In human dialogue, clarification questions are a routine, low-cost repair mechanism \citep{purver2004theory, chiyahgarcia2023referring}, but one used {\it strategically} rather than by default.

When compared to humans under similar conditions of ambiguous referents, LLMs produce fewer grounding acts and instead presume common ground \citep{shaikh2024grounding}. Faced with ambiguous questions, LMs ask for clarification less often than humans do, answering with one interpretation instead \citep{kuhn2023clam}. Studies have shown that LLM-based chat systems can guess in response to ambiguous queries \citep{deng2023prompting}. In multi-turn settings, assumptions made early in the dialogue lead to misunderstandings and errors \citep{laban2026lost}. Given referentially ambiguous questions, this can surface as erroneously committing to one candidate object \citep{testoni2025racquet}.

However, not every input calls for a question \citep{zhang2025clarify}, and each question costs the user time and effort; a good policy should do so only when necessary and ask questions when the answer meaningfully changes the decision \citep{rao2018learning, aliannejadi2019asking}. The question's content must follow the uncertainty \citep{testoni2024asking}.

In this paper, we examine the problem of ``strategic clarification''---asking clarification questions only when the answer will change the result. We construct a controlled testbed made of episodes of the SIMMC~2.1 benchmark \citep{kottur2021simmc, simmc21}, and examine when common LLM optimization techniques successfully decide to ask questions based on whether the answer will reduce the surviving set of remaining candidates in a reference game. We also propose \textbf{BACE (Belief-Aware Critic with Epistemic tracking)}, a method that uses RL to train a lightweight action chooser on top of a frozen language model, and show that BACE outperforms GRPO \citep{shao2024deepseekmath} and ArCHer \citep{zhou2024archer} on choosing the right number of clarifying questions to optimally isolate the referent.\footnote{Code and data: \url{https://github.com/csu-signal/Clarifying_Ambiguous_References}}

\vspace*{-2mm}
\section{Related Work}
\label{sec:related}
\vspace*{-2mm}

Clarification is a key component of building common ground through shared referencing \citep{clark1986referring, clark1991grounding}. \citet{paek2000conversation} cast it as a decision under uncertainty: keep a probability over the user's intention and choose between acting and asking by expected utility. LLMs produce fewer of these grounding acts than people \citep{shaikh2024grounding}, possibly because a label given to a single reply can't credit a question for what its answer enables \citep{herlihy2024overcoming, zhang2025modeling}. 

Work on clarification separates deciding whether a request needs clarifying from deciding which question to pose \citep{aliannejadi2021building, zhang2025clarify}. The first is usually scored as classification against a graded or binary ambiguity label \citep{aliannejadi2021building, zhang2024clamber} or by the gain in the final answer when only a fixed fraction of inputs may be clarified \citep{zhang2025clarify}, and \citet{kim2024aligning} align a model to ask where its own attempt to disambiguate gains the most information. The second is judged against reference questions or by its effect downstream \citep{rao2018learning, aliannejadi2019asking, zhang2024clamber}. CLAM \citep{kuhn2023clam} scores the final answer instead, discounting correct answers to unambiguous questions on which the model still asked. 

In GuessWhat?!, a questioner locates an object in an image through yes/no questions \citep{devries2017guesswhat}. PhotoBook \citep{haber2019photobook} accounts for shared information accumulated over a chain of references. SIMMC~2.0 \citep{kottur2021simmc} sets shopping dialogues in photo-realistic scenes and labels whether a turn is ambiguous and which objects each mention resolves to, and SIMMC~2.1 \citep{simmc21} adds object annotations to ambiguous mentions. \citet{chiyahgarcia2023referring} test whether models can interpret clarification exchanges already in SIMMC dialogues.

ArCHer \citep{zhou2024archer}
pairs an off-policy utterance-level critic with a token-level actor. Agentic work runs GRPO \citep{shao2024deepseekmath} or PPO over full trajectories \citep{jin2025search}, where RAGEN \cite{wang2025ragen} observes reward variance cliffs and gradient spikes. STaR-GATE \citep{andukuri2024stargate} fine-tunes on questions that make a good response likelier, and CollabLLM \citep{wu2025collabllm} rewards replies by their effects on simulated future turns. ACT \citep{chen2025learning} is closest to our decision: it builds DPO pairs that contrast clarifying with answering, on data where only some turns need a question. 

A learned scorer can improve a generator without changing its weights by choosing among its samples \citep{cobbe2021training,verma2022chai}; \citet{zhou2024archer} argue that this slows policy improvement. POMDP dialogue managers act on a distribution over the dialogue state \citep{williams2007pomdp, young2013pomdp}, and several methods pick the question with the highest expected information gain at inference time, without training \citep{lee2018answerer, hu2024uncertainty, kobalczyk2025active}, while \citet{deng2026uncertainty} reward a trained clarifier by how far an exchange moves a Bayesian belief toward the user's goal.

This work addresses all the above questions by decoupling the action choosing and response generation components in a controlled environment that enables assigning credit to different model capabilities.
In our environment, credit given to a question is exactly computable: a question costs $\lambda$ when it is asked, and its value shows up in the reward of the answer it leads to.
We use the SIMMC~2.1 candidate annotations, but our policy decides whether to ask, what to ask about, and when to commit to a referent.
Within one episode, we score whether a policy asks exactly on the ambiguous turns, and whether it stops after as many questions as an optimal oracle predicts. The candidates are listed, replies and rewards follow from the annotations, and gold depth gives the questions each episode needs, so stopping is scored along with asking.
In BACE, information gain is neither the objective nor the reward: the chooser reads the belief, as a list of the remaining candidates in its prompt and as features that include each question's expected information gain, and learns from the task reward when asking pays off.

\vspace*{-2mm}
\section{Task and Environment}
\label{sec:data}
\vspace*{-2mm}
\subsection{Task}
\label{ssec:task}
\vspace*{-2mm}

\nk{The central task we address is one of reference disambiguation via dialogue. Given a target item whose initial description admits many possible candidates, two agents---a \textit{user} and an \textit{assistant}---engage in a \textit{reference game} \citep{clark1986referring} where the assistant tries to narrow the candidate set down to a single item in as few turns as possible. At each turn the assistant must decide whether to \textsc{Ask} a question about one of the item's attributes, or to \textsc{Answer} by naming a specific candidate and ending the episode.}

\vspace*{-2mm}
\subsection{Environment}
\label{ssec:env}
\vspace*{-2mm}

\nk{Our environment is constructed from the SIMMC~2.1 dataset \citep{kottur2021simmc, simmc21}. }
An episode is a user turn from SIMMC~2.1 that refers to an object (Fig.~\ref{fig:task}; Fig.~\ref{fig:task-example} in Appendix~\ref{app:examples}). The policy can observe the dialogue history, the user's utterance, and the list of candidate objects with their non-visual attributes (brand, price, customer review, available sizes, size) and position in the scene and the relations between objects.\footnote{Visual attributes such as color, pattern and type are withheld, following SIMMC's rule of forbidding visual metadata at inference time \citep{simmc2inputs}.} At each turn the policy can either \textsc{Answer} or \textsc{Ask}. \textsc{Ask} poses a question about one of the eleven askable attributes (brand, price, customer review, available sizes, size, left/right and top/bottom position within the scene, or relative location to another remaining candidate---in terms of four cardinal directions). The environment replies as a deterministic scripted user, with the true answer to the question. This question--reply pair is appended to the observation before the next decision, so the policy can rule out candidates whose values don't match. We call the remaining candidates the {\it survivor set}.

A correct answer earns $+1$; a wrong one, $-1$. Each question incurs a cost of $\lambda = 0.1$. The policy may ask at most 3 questions; further ones are refused (Appendix~\ref{app:dataset}). The mean reward over a batch of turns is therefore a fixed function of accuracy and the average number of questions ($\bar{q}$):
\vspace*{-2mm}
\begin{equation}
  \bar{r} = (2 \cdot \mathrm{acc} - 1) - \lambda \cdot \bar{q}.
  \label{eq:reward}
\vspace*{-2mm}
\end{equation}
A policy that never asks pays nothing in questions but guesses on every ambiguous turn, where a uniform guess among $n$ candidates is right with probability $1/n$. A policy that always asks pays for questions it didn't need, so the reward favors asking only when the reply can change the answer.

The environment tracks the survivor set but the policy never sees it; it keeps the initial candidate list throughout and has to reason out what has been ruled out from the observed questions and replies. A question that can't be mapped to a known attribute, or that repeats an attribute already asked, reveals nothing but uses up a turn.

\vspace*{-2mm}
\subsubsection{Episode Sampling}
\label{ssec:episodes}
\vspace*{-1mm}

We use the pre-defined splits from SIMMC~2.1: {\tt train}, {\tt dev}, and {\tt devtest}~\citep{kottur2021simmc, simmc21}.\footnote{The public release of the {\tt test-std} split carries no annotations, so no episodes can be built from it; we use {\tt dev} during development and {\tt devtest} as the test set.} We construct episodes from training through \textbf{sampling from SIMMC~2.1} and through \textbf{synthetic data generation}.

\paragraph{Episodes from SIMMC 2.1}

If SIMMC~2.1 marks a mention as unambiguous, the single referent is the only candidate and the correct action is to answer. No further clarification questions are required. For mentions marked ambiguous, SIMMC~2.1 annotates every object in the scene the user could mean as of the end of the dialogue history. We use these as the candidate list $\mathcal{C}$. 
The dialogue continues from this starting point. If two or more candidates remain in the survivor set after a question and response, the policy cannot know the referent without asking.

We find the gold referent by looking ahead to later turns in the dialogue for the first subsequent turn in which the referenced objects form a strict subset of the candidates, $\mathcal{C}^\star \subset \mathcal{C}$. When $|\mathcal{C}^\star| = 1$ the single object is extracted as the ground truth target (this occurs in 48\% of ambiguous training episodes). When $|\mathcal{C}^\star| = 2$, we take the first one (29\%). When no later turn narrows the set ($\mathcal{C}^\star = \mathcal{C}$), we pick one candidate at random with a fixed seed (23\%). The attribute questions that separate the candidates come from the scene annotations, so these fallbacks don't change which questions are useful, only which object is considered correct. The shares for each split are in Appendix~\ref{app:dataset}.

Each episode also has a {\it gold depth}: the optimal number of further questions needed for referent resolution. To compute this, an oracle that knows the referent repeatedly asks about the attribute that best splits the remaining survivors, each attribute at most once, until only the referent is left. The number of questions asked is the episode's gold depth; unambiguous episodes have depth 0. Gold depth is a property of the episode, not of the policy, so it splits every method's results identically. The few episodes the available attributes can't resolve stay in evaluation and out of training (Appendix~\ref{app:dataset}). Table~\ref{tab:data} shows the resulting distribution.

\begin{table}[t]
  \centering
  \footnotesize
  \setlength{\tabcolsep}{3pt}
  \begin{tabular}{@{}lrrrrrrr@{}}
    \toprule
    \textbf{Depth} & \textbf{0} & \textbf{1} & \textbf{2} & \textbf{3} & \textbf{4--5} & \textbf{Unres.} & \textbf{Total} \\
    \midrule
    Train & 7,405 & 2,613 & 840 & 162 & 15 & 47 & 11,082 \\
    Dev & 617 & 247 & 74 & 18 & 3 & 1 & 960 \\
    Devtest & 1,701 & 550 & 184 & 33 & 3 & 12 & 2,483 \\
    Synth. & -- & 4,792 & 6,565 & 7,243 & -- & -- & 18,600 \\
    \bottomrule
  \end{tabular}
\vspace*{-2mm}
  \caption{Episodes per gold depth, the number of questions the oracle needs (Unres.: unresolvable). Synth.: synthetic episodes added to the real training split.}
  \label{tab:data}
\vspace*{-2mm}
\end{table}

\vspace*{-2mm}
\subsection{Anchored Synthetic Episodes}
\label{sec:synthetic}
\vspace*{-2mm}

Two-thirds of the training episodes need no question and a quarter need exactly one, while only 162 of 11,082 need three (Table~\ref{tab:data}), leaving very few examples of multi-question behavior. To fill the multi-question buckets, we generated synthetic training episodes from real ambiguous ones. Each synthetic episode starts from a template---a real ambiguous turn with at least three annotated candidates. We keep the template's scene, dialogue history and utterance, replace its candidate list with a strict subset of at least two of the annotated candidates, and pick a gold referent from that subset. Since SIMMC~2.1 annotates every candidate as a possible referent of the utterance, any of them is a valid gold referent, and a smaller set with a different referent can need more questions than the original. Only the choice of candidates and referent is new; no text or attribute value is invented. We then run the oracle of Sec.~\ref{ssec:episodes} to find its gold depth.

We discard synthetic episodes whose utterance no longer fits the smaller candidate set, duplicates, and any beyond 20 per template and depth (Appendix~\ref{app:synthetic}). We generate episodes at depths 1 to 3 until each of them---real and synthetic episodes together---is as large as the 7,405 real episodes at depth 0; no real episode is removed or re-weighted. A bag-of-n-grams classifier separates synthetic from real episodes of the same depth with a held-out AUC of 0.59--0.63, against 0.50--0.57 for two random halves of the real ones.

Synthetic episodes are used only to train the SFT checkpoint, which is also the starting point of every RL method in Sec.~\ref{sec:baselines}. The RL methods train on real episodes, and every number we report is on the real {\tt dev} and {\tt devtest} splits.

\vspace*{-2mm}
\section{BACE (Belief-Aware Critic with Epistemic tracking)}
\label{sec:bace}
\vspace*{-2mm}

The baseline expressivity of LLMs already gives them the ability to write an appropriate clarifying question in principle; however, a good question at the wrong time may confound reference resolution by focusing on irrelevant attributes, resulting in a response that contributes no new information. A model must first decide if asking a question is even the appropriate move at the current timestep. If one model both decides and writes, correcting a wrong decision to ask or answer also changes the weights that write the question or response.

\begin{figure}[!t]
  \centering
  \includegraphics[width=.7\columnwidth]{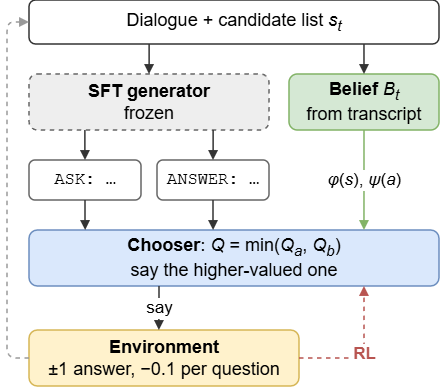}
\vspace*{-2mm}
  \caption{BACE at one turn. The frozen SFT model writes a candidate question and a candidate answer. The chooser scores both from the backbone's hidden state and features of a belief $B$ over the candidates, built from the questions and replies so far, and picks the one with the higher value. The environment's reward trains only the chooser.}
  \label{fig:bace}
\vspace*{-2mm}
\end{figure}

Instead, given a belief over sets of candidate referents, a candidate question, and a candidate answer, a simple chooser can decide on the right action given the belief state, and the generation on that branch can be inserted into the dialogue. {\bf BACE (Belief-Aware Critic with Epistemic tracking)} separates deciding from writing with a simple intuition: the same quality of question can improve outcomes just by asking at the right time. The frozen SFT model writes both candidates, and RL trains only the chooser (Fig.~\ref{fig:bace}).

\vspace*{-2mm}
\subsection{Belief}
\label{sec:bace-belief}
\vspace*{-2mm}

Since the policy cannot see the survivor set (Sec.~\ref{ssec:env}), BACE tracks a belief $B$: the candidates from the initial list left open by the replies so far. $B$ starts as the full candidate list, and each reply narrows it by the environment's rule for the survivor set, using only the attributes, positions and relations in the observation (Appendix~\ref{app:belief}). The result tracks the survivor set closely: when we replay every episode with random question orders, $B$ equals the survivor set in all but 83 to 95 of 348,600 states. We also extract seven belief-related features from the questions and replies so far, which both candidates share (Eq.~\ref{eq:bace-phi}):

\vspace*{-2mm}
\begin{align}
  \varphi(B) = \notag\big[\, &b/n,\, \log_2 b/\log_2 n,\, 1/b,\, k/K,\,\\
  &1{-}k/K,\, \mathbb{1}[b{=}1],\, |A|/11 \,\big],
  \label{eq:bace-phi}
\end{align}
\vspace*{-2mm}

\noindent
where $b = |B|$, $n$ is the number of candidates, $k$ the questions asked so far out of the budget $K = 3$ (refused ones included), and $A$ the set of distinct attributes asked so far, out of the 11 askable ones.

\vspace*{-2mm}
\subsection{Generator and Chooser}
\label{sec:bace-chooser}
\vspace*{-2mm}

BACE's prompt adds one line to the observation, listing the candidates still in $B$ (Sec.~\ref{sec:bace-belief}), e.g., \texttt{Still possible (2): [46, 49]}; Appendix~\ref{app:bace-ablations} trains without it. The features go to the chooser's heads, not into the prompt.
The response generator is a simple frozen SFT expert model (Sec.~\ref{sec:sft}). At every turn it samples completions after \texttt{ASK:} and after \texttt{ANSWER:} and keeps one of each as a finalist (Table~\ref{tab:hp-bace}). Every question and answer BACE produces is therefore written by the SFT model. Training only changes when it asks.

The chooser is inspired by ArCHer's critic \citep{zhou2024archer}: its own LoRA adapter on the backbone, with two pairs of Q and V heads. To score a finalist, we run the backbone with this adapter on the observation followed by the finalist's (\texttt{ASK:~<attribute>~|~<question>} or \texttt{ANSWER:~<id>}) description, and take the last layer's hidden state at the final token. A Q head maps that vector, together with the belief features (Sec.~\ref{sec:bace-belief}), to the finalist's value. A V head does the same from the observation alone, with only the features shared by both finalists. The value of a finalist is the smaller of its two Q values. At evaluation the chooser picks the finalist with the higher value. During training it samples between them, so that both branches keep being tried (Appendix~\ref{app:hyper}).

Each finalist also gets three features of its own, $\psi$ (Appendix~\ref{app:belief}): for a question, the entropy its attribute is expected to leave in $B$ and whether it was already asked; for an answer, whether the named object is still in $B$ and the chance $1/b$ that a uniform guess from $B$ is right.

\vspace*{-2mm}
\subsection{Reward and Critic}
\label{sec:bace-critic}
\vspace*{-2mm}

The chooser is trained on the task reward: $\pm1$ for the answer and $-\lambda$ per question. Training uses a target critic, a copy of the critic that stays fixed during an RL iteration and is refreshed at the end of it. Each Q head is fit by squared error to the average of two estimates of the transition's value:
\begin{equation}
  y_Q = \tfrac12\big(r + \bar V(s')\big) + \tfrac12\,G .
  \label{eq:bace-q}
\end{equation}
The first term is the one-step TD target: the reward $r$ of the transition plus the target critic's value $\bar V$ of the next state. The second is scaled $G$, the return actually observed from this transition to the end of the episode. At the last step, $y_Q = r$. Each V head is fit to the target critic's value of the greedy choice (Appendix~\ref{app:hyper}).

\vspace*{-2mm}
\section{Training and Experiments}
\label{sec:baselines}
\vspace*{-2mm}

Every method we evaluate uses Qwen2.5-7B-Instruct \citep{qwen2025qwen25} as its backbone, with post-training using LoRA adapters \citep{hu2021lora} on the frozen weights. All methods receive the same observation (Sec.~\ref{ssec:env}). Trained baselines output their decisions as a single line, either \texttt{ANSWER:~<id>} or \texttt{ASK:~<attribute>~|~<question>}, with the attribute copied from the list of askable ones. Reinforcement learning (RL) baselines all start from the same SFT checkpoint (Sec.~\ref{sec:sft}), with a fresh adapter and a KL penalty to the SFT policy. The RL methods train on the real training episodes only. Appendix~\ref{app:hyper} gives the hyperparameters of every method and Appendix~\ref{app:prompts} the prompts. Prompts for each model are calibrated to the requirements on their inputs.

\vspace*{-2mm}
\subsection{Natural Behavior} 
\label{sec:natural}
\vspace*{-2mm}

Natural Behavior is the instruct model with no training and no task description. Its prompt (Fig.~\ref{fig:prompt-instr}) says it is a shopping assistant, gives the observation, including the list of candidate objects with their IDs, and asks it to respond to the user; it is not told that it has to identify the referent, that it may ask, or how to format an answer. This demonstrates default model behavior: deciding by itself whether a turn needs a question. Its replies are turned into questions and answers by the rules in Appendix~\ref{app:prompts}, and Appendix~\ref{app:prompted} compares it with five prompted policies on the same backbone.

\vspace*{-2mm}
\subsection{Supervised Fine-Tuning}
\label{sec:sft}
\vspace*{-2mm}

SFT imitates the oracle that defines gold depth (Sec.~\ref{ssec:episodes}). For each training episode we follow the oracle's chain and make one example per turn, pairing the observation at that turn with the oracle's decision. Once one survivor is left, the target is \textsc{Answer} with that object. The question text comes from a fixed template for each attribute, because SFT is meant to learn the decision and the attribute rather than the wording. We train for one epoch on the real and synthetic episodes (Sec.~\ref{sec:synthetic}).

SFT learns which decision the oracle makes in each state and nothing about what that decision is worth. Its loss counts an unneeded question and a wrong answer the same way, as a mismatch with the oracle. Imitation can't weigh one mistake against another \citep{ross2014reinforcement}, so SFT's decisions don't change with the cost of a question. It also needs the oracle's action at every state, and its accuracy on multi-question episodes grows with the number of oracle-labelled episodes (Table~\ref{tab:sft-pool}). RL needs only the outcome of each episode, and can beat imitation when a few states decide that outcome \citep{kumar2022should}, as the decision to stop does here. We start every RL method from SFT and report it as a reference.

\vspace*{-2mm}
\subsection{GRPO}
\label{sec:grpo}
\vspace*{-2mm}

GRPO \citep{shao2024deepseekmath} replaces the value network with a group baseline: sampling several responses to the same input and scoring each against the group mean. We use trajectory-level GRPO, which extends GRPO to multi-turn tasks by treating the whole interaction as one completion \citep{wang2025ragen, jin2025search}. For each training episode we sample $K = 4$ full rollouts from the current policy at temperature 0.8, sampling again at every turn. A rollout's return is its episode reward, and its advantage is that return normalized by the mean and standard deviation of the group's returns. The same advantage is applied to every token the policy generated at every turn of the rollout, with PPO's clipped surrogate objective \citep{schulman2017proximal}. There is no critic and no credit per turn, so a rollout that asked one useful question and one redundant one gives both the same advantage. A group whose rollouts all get the same return has zero advantage and gives no update. We train on one pass over the 11,035 real training episodes.

\vspace*{-2mm}
\subsection{ArCHer}
\label{sec:archer}
\vspace*{-2mm}

ArCHer \citep{zhou2024archer} splits multi-turn RL into an utterance-level critic, trained off-policy with temporal-difference updates from a replay buffer, and a token-level actor trained by policy gradient on the critic's advantage. We follow its critic: two pairs of Q and V heads on the backbone with a separate LoRA adapter (Table~\ref{tab:hp-archer}). In our adaptation the critic also makes the ask/answer decision. At every turn the actor writes both branches, the critic scores them and picks the one with the higher Q value, and the actor's output after the chosen prefix, \texttt{ASK:} or \texttt{ANSWER:}, is what gets said. 

As in the original, the first 10 of the 28 iterations train only the critic, and we made two changes for a 7B actor: a KL penalty to the SFT policy and a question attribute taken from the actor's probabilities (Appendix~\ref{app:hyper}). A run makes one pass over the real training episodes, 28 iterations of 394 (11,032).

\vspace*{-2mm}
\subsection{BACE Training}
\label{sec:bace-training}
\vspace*{-2mm}

With untrained heads the chooser picks at random, so before RL we fit it to returns measured at each state. We take 150 training episodes and, at every state $s$ along them, clone the environment and play out each branch twice: answering and ending the episode, and asking and then giving one greedy answer. The two branches share their random seeds, so the difference comes from the action rather than from sampling. Averaging the two playouts gives $G_{\text{ask}}(s)$, which includes the question's cost $\lambda$, and $G_{\text{ans}}(s)$. With $a \in \{\text{ask}, \text{ans}\}$, $\Delta Q_i = Q_i(s,\text{ask}) - Q_i(s,\text{ans})$, $\Delta G = G_{\text{ask}} - G_{\text{ans}}$ and $\bar{G} = \tfrac12 (G_{\text{ask}} + G_{\text{ans}})$, the warm start minimizes $\mathcal{L}(s)$ over six epochs of the states:

\vspace*{-4mm}
\begin{equation}
\begin{aligned}
  \mathcal{L}(s) &= \sum_{i=1}^{2} \Big[ \sum_{a} \big(Q_i(s,a) - G_a\big)^2 \\
  &\quad + \big(\Delta Q_i - \Delta G\big)^2 + \big(V_i(s) - \bar{G}\big)^2 \Big].
\end{aligned}
\label{eq:bace-warm}
\vspace*{-2mm}
\end{equation}

The third term fits the gap between the branches directly, since the gap is what the chooser acts on; with only the first two, a head can fit the level both returns share and leave the gap near zero. Measuring both branches at every state gives each action its own return; returns from SFT's own rollouts would mostly follow answers. Because two thirds of states need no question, we also resample the states so that $|B| > 1$ in half of them.

RL then runs through one full pass over the real training episodes, the same data GRPO and ArCHer see, in 552 iterations. Each one rolls out 20 episodes under the chooser into a replay buffer and trains the critic on transitions drawn from it (Table~\ref{tab:hp-bace}). A run trains 27.7M parameters; GRPO trains 40.4M (Table~\ref{tab:cost}).

\vspace*{-2mm}
\section{Evaluation}
\label{sec:eval}
\vspace*{-2mm}

\subsection{Populations}
\label{sec:eval-pop}
\vspace*{-2mm}

Each evaluation covers a whole split: the 960 real {\tt dev} episodes or the 2,483 real {\tt devtest} episodes (Table~\ref{tab:data}).
Policies act in the environment of Sec.~\ref{ssec:env} with a budget of three questions and $\lambda = 0.1$. Decoding is greedy, apart from the candidate questions and answers that BACE samples (Sec.~\ref{sec:bace-chooser}). This sampling is seeded from the episode's identifier, so all checkpoints share the same seed on a given episode.

Episodes are grouped by the number of questions the oracle of Sec.~\ref{ssec:episodes} asks under that budget: ask-0 holds the unambiguous episodes, and ask-1 to ask-3 the ambiguous ones. Ask-3 thus includes the episodes that need more than three questions or cannot be resolved at all (Table~\ref{tab:data}). Since an unambiguous episode lists only its referent, the list tells a policy whether to ask. Ask-1 thus mainly tests the choice of question, and whether a policy has learned when to stop shows only on ask-2 and ask-3.

\vspace*{-2mm}
\subsection{Metrics}
\label{sec:eval-metrics}
\vspace*{-2mm}

All methods are scored with the metrics of Appendix~\ref{app:results} on these populations. Each trained RL method is run on seeds 0, 1, 7, 17 and 123, and we report the mean and standard deviation over them. The untrained policies decode greedily and are deterministic, so they are single runs. We report each method at the end of one full pass over the train split, so all of them have seen the same episodes. The results and figures of this section and the next are on devtest; Appendix~\ref{app:results} tabulates them, and Appendix~\ref{app:dev} repeats them on dev.

\begin{figure*}[!t]
  \centering
  \includegraphics[width=0.18\textwidth]{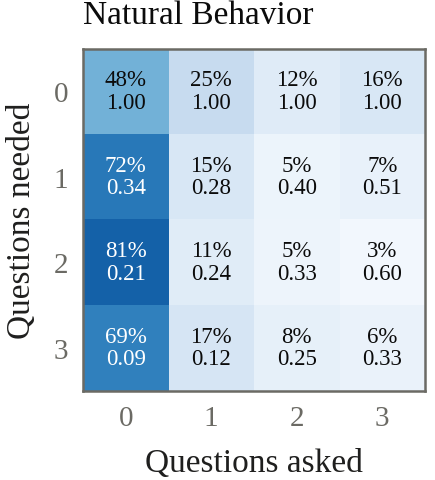}\hfill
  \includegraphics[width=0.18\textwidth]{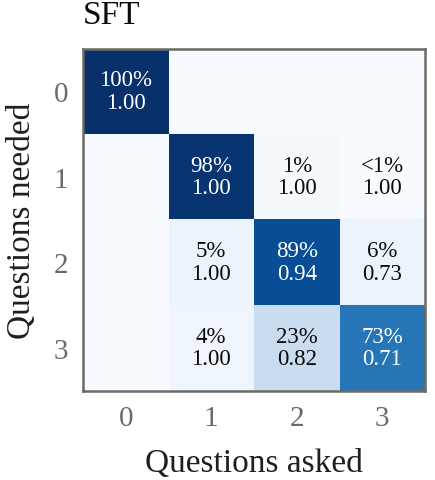}\hfill
  \includegraphics[width=0.18\textwidth]{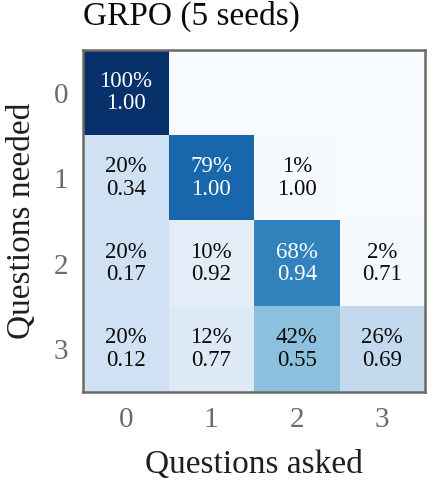}\hfill
  \includegraphics[width=0.18\textwidth]{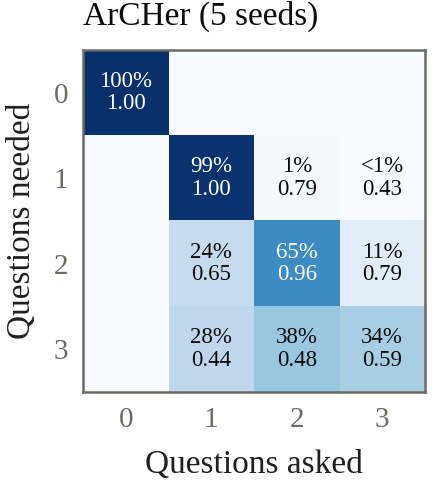}\hfill
  \includegraphics[width=0.18\textwidth]{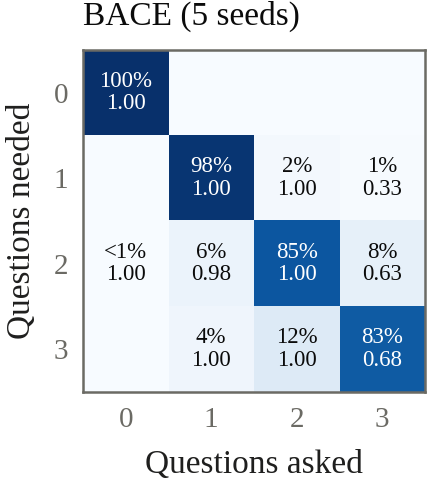}
\vspace*{-2mm}
  \caption{Questions the oracle needs (rows) against questions the policy asked (columns) on {\tt devtest}, at the points of Table~\ref{tab:main}, with the seeds of a method pooled. Each cell gives its share of the row and the accuracy of the episodes in it. The diagonal is a hit; cells left of it answer early and cells right of it answer late.}
  \label{fig:depth-asks}
\vspace*{-2mm}
\end{figure*}

\begin{figure*}[!t]
  \centering
  \resizebox{0.75\textwidth}{!}{%
    \begin{minipage}{\textwidth}
      \centering
      \includegraphics[width=0.28\textwidth]{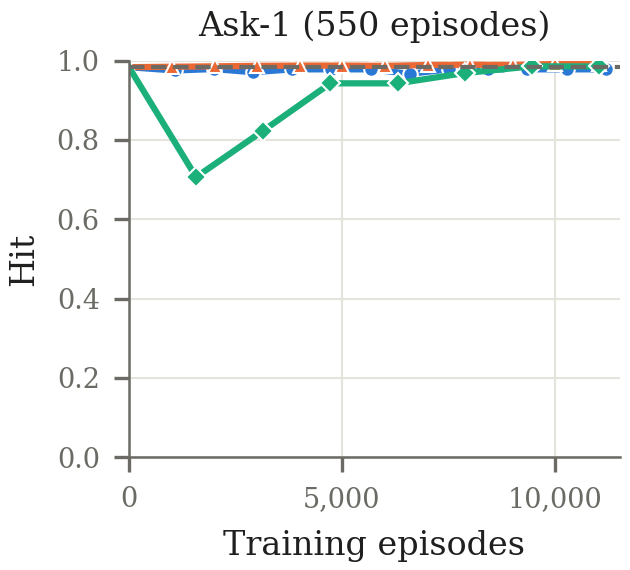}\hfill
      \includegraphics[width=0.28\textwidth]{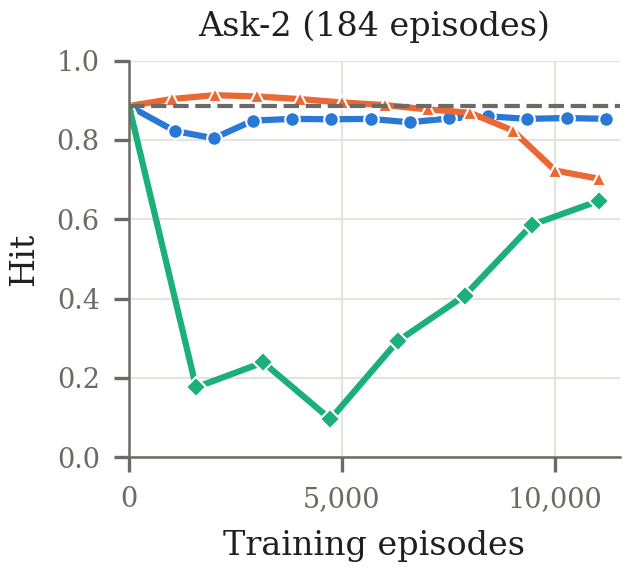}\hfill
      \includegraphics[width=0.28\textwidth]{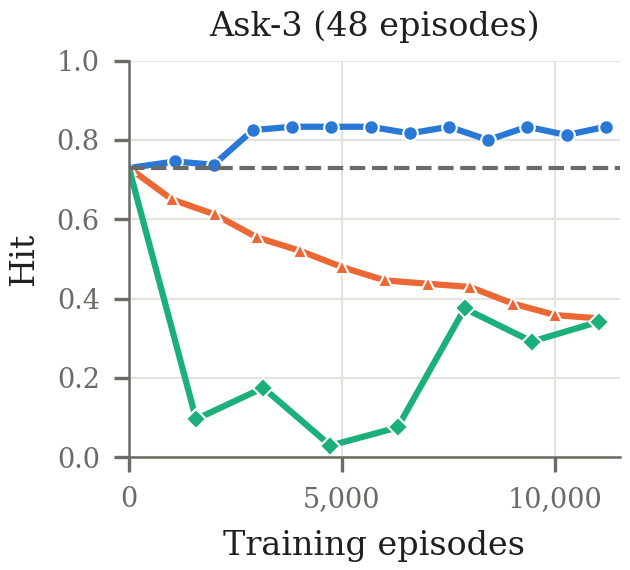}\\[2pt]
      \includegraphics[width=0.28\textwidth]{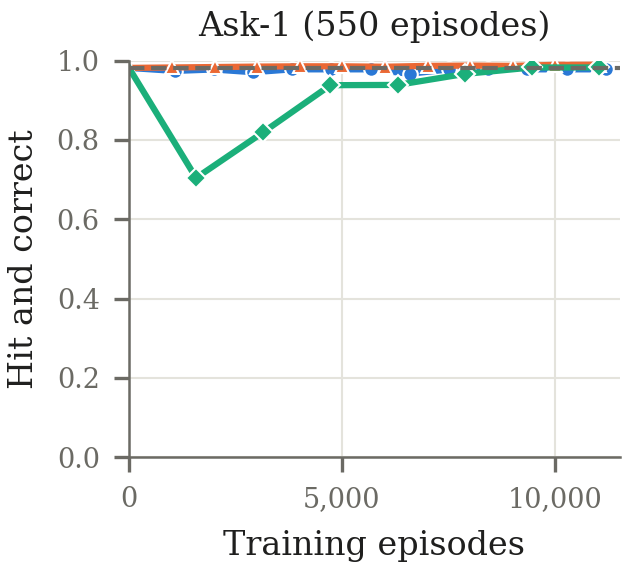}\hfill
      \includegraphics[width=0.28\textwidth]{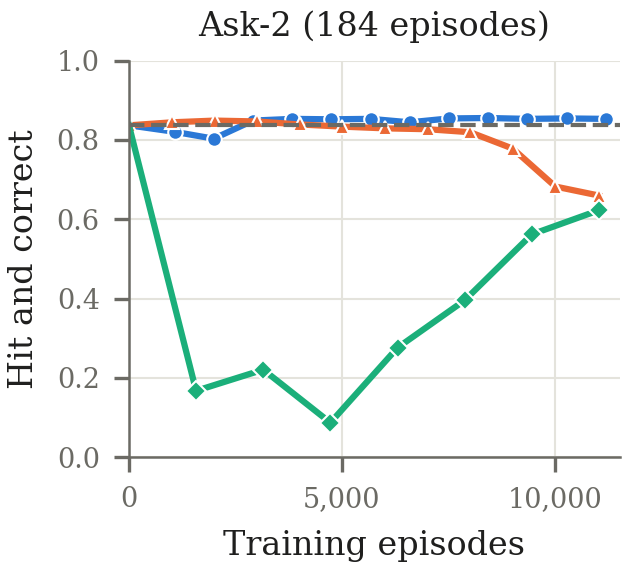}\hfill
      \includegraphics[width=0.28\textwidth]{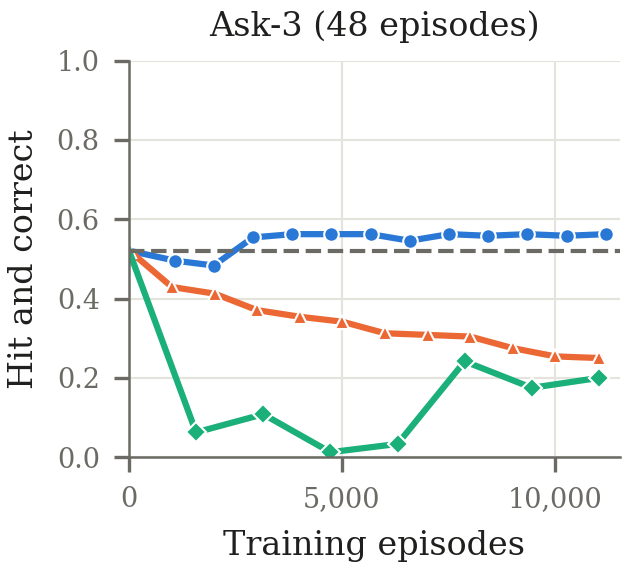}\\[2pt]
      \includegraphics[width=0.75\textwidth]{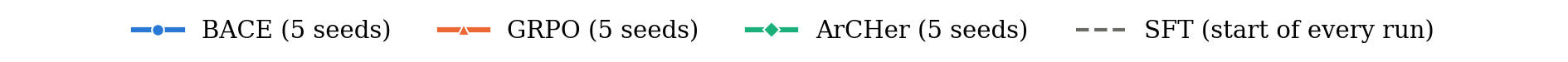}
    \end{minipage}%
  }
\vspace*{-2mm}
  \caption{Hit (top) and hit-and-correct (bottom) on {\tt devtest} against training episodes, by the number of questions the oracle needs. A hit answers after exactly as many questions as the oracle; a hit-and-correct also ends on the correct referent. Ask-3 includes episodes the oracle can't resolve within the budget.}
  \label{fig:curves-hit}
\vspace*{-2mm}
\end{figure*}

The metrics cover accuracy and cost, the decision to ask, and the decision to stop (Table~\ref{tab:metrics}). Reward is the mean return (Eq.~\ref{eq:reward}). Ask F1 scores whether a question was asked against whether the episode is ambiguous, and hallucinated certainty is the share of ambiguous episodes answered wrongly without a question. A hit answers after exactly as many questions as the oracle, and a hit-and-correct also ends on the correct referent. Results are given on the full split and by ask-$k$ group. A matrix of questions needed against questions asked shows how often a policy answers early, on time or late, and how accurate it is in each case.

\vspace*{-2mm}
\section{Results and Discussion}
\label{sec:results}
\vspace*{-2mm}

We compare the methods at the end of one pass over the train split, on the full split and by the number of questions the oracle needs (Tables~\ref{tab:main} and~\ref{tab:groups}, Appendix~\ref{app:results}). Figure~\ref{fig:depth-asks} crosses the questions needed with those asked, and Figure~\ref{fig:curves-hit} follows each method through the pass.

\vspace*{-2mm}
\subsection{Main Comparison}
\label{sec:res-main}
\vspace*{-2mm}

Left to itself, the instruct model can't tell when a question is needed. Natural Behavior asks on 51.9\% of the episodes that need no question and leaves 73.9\% of the ambiguous ones without one (Ask F1 0.218). Its accuracy (0.782) is no higher than that of Direct Answer, which never asks (0.792), so its questions only cost it reward: 0.486 against 0.584. Few of them are clarification: most are offers to add an item to the cart or help with something else, and only a handful ask which object the user means (Appendix~\ref{app:examples}). 

SFT learns the decision from the oracle. It asks on every ambiguous episode and on no unambiguous one, stops after the oracle's number of questions on 98\% of episodes, and earns 0.938 devtest reward.
GRPO and ArCHer start from this checkpoint, and neither improves on it. GRPO keeps asking on every seed, but on seed 1 it stops asking a second question after about 8,000 training episodes (Fig.~\ref{fig:curves-hit}); by the end it answers after one question on 84\% of ask-2 and ask-3 episodes and reaches 0.881 reward, which accounts for most of its SD of 0.025. ArCHer stays close to SFT on every seed (0.918 $\pm$ 0.011 reward) but falls behind on ask-2 and ask-3 episodes, where it answers after the right number of questions on 64.7\% and 34.2\% of them, against SFT's 89.1\% and 66.7\%. Most of its misses there are early answers, usually one question short (Fig.~\ref{fig:depth-asks}). Its seeds end within 0.027 reward of each other, and none of them reaches SFT's hit on these episodes (at best 78.8\% and 58.3\%).

BACE ends the pass at 0.940 $\pm$ 0.000 reward, the only RL result at or above SFT. The warm start alone gives 0.881 $\pm$ 0.064 reward, below SFT, and RL raises every seed from there (Figure~\ref{fig:curves-hit}). The five seeds agree on 2,481 of the 2,483 episodes, and on every seed over 99\% of the decisions follow one rule: ask while more than one candidate is left in $B$. On ask-2 and ask-3 episodes it is well ahead of both RL methods (hit-and-correct 0.853 and 0.562, against 0.660 and 0.250 for GRPO and 0.623 and 0.200 for ArCHer). Against SFT it answers on time more often on ask-3 episodes (hit 0.833 against 0.667) and a little less often on ask-2 (0.853 against 0.891), and it answers on time and correctly about as often on ask-2 episodes (hit-and-correct 0.853 against 0.859) and more often on ask-3 (0.562 against 0.479). 

\vspace*{-2mm}
\subsection{Discussion}
\label{sec:discussion}
\vspace*{-2mm}

How good the counterfactual warm start is varies from run to run (0.772 to 0.932 devtest reward), and the full pass brings all five seeds to 0.940 or 0.941, while shorter runs can lose the decision as the chooser's values for \textsc{Ask} and \textsc{Answer} converge on ambiguous states (Appendices~\ref{app:bace-seeds} and~\ref{app:bace-stability}).

SFT is within 0.002 of BACE's reward in Table~\ref{tab:main}, and the oracle it imitates is close to optimal for this reward by construction, but it needs the gold referent and the survivor set at every state, while RL only needs the outcome. BACE's questions stay close to SFT's (Appendix~\ref{app:questions}), so where the two differ, on ask-2 and ask-3 episodes, it is in when the chooser stops, which it decides from $\varphi$ and $\psi$; the \texttt{Still possible} line speeds up training. With SFT trained on real episodes only as the generator (0.908 by itself), BACE reaches 0.945 after 2,000 episodes (Appendix~\ref{app:bace-ablations}).

\vspace*{-2mm}
\section{Conclusion}
\vspace*{-2mm}

In this paper, we studied strategic clarification: asking a clarifying question only when its answer will narrow the set of candidate referents. We did this through a SIMMC~2.1 reference game that records how many questions each ambiguous reference needs. Left to itself, the instruct model asks on the wrong episodes and earns less reward than never asking. Imitating an oracle teaches it when to ask, but GRPO and ArCHer fall behind imitation on episodes that need several questions. BACE freezes the imitation model and trains only a small chooser that picks between asking and answering from a belief over the remaining candidates. It is the only RL method to match SFT, and on episodes that need several questions it answers on time and correctly more often than GRPO and ArCHer.

Future work may exploit the visual modality in SIMMC and other datasets. Vision-language models faced with an ambiguous reference to an image tend to commit to the most salient object \citep{testoni2025racquet}, and we have not tested whether this holds here, or whether GRPO, ArCHer and BACE keep their ranking once visual questions can be asked and each candidate has to be found in the image.

\section*{Limitations}

The main limitation of this work is that it is text-only. Our test environment constructed from SIMMC~2.1 serves as a controlled scenario that allows us to isolate the effects of settings such as the warm start and training stability across candidate set sizes. We do this to avoid introducing additional stochasticity, such as from the outputs of a vision encoder. 

Our results also come from a single dataset. The oracle and the SFT targets rely on its annotation of every object an ambiguous mention could refer to, so moving to other domains \citep{khebour2024common, haber2019photobook} would first need a model to propose the candidates.

In our experiments the policies are also handed the candidate list, which makes deciding whether to ask easier than for a system that has to find the candidates itself. Finally, the scripted user always answers exactly, one attribute per question, whereas real users can be vague or wrong, and we have yet to evaluate policies with such human-like behavior that can demonstrate the robustness of our work.

\section*{Ethics Statement}

We use SIMMC~2.1 \citep{simmc21}, released under CC BY-NC-SA 4.0, and Qwen2.5-7B-Instruct \citep{qwen2025qwen25}, released under Apache 2.0. Our use is non-commercial research, which both licenses permit. The real and synthetic episodes we derive from SIMMC~2.1 are adaptations of it, so we release them under the same CC BY-NC-SA 4.0 license, for research use only.

We used Claude Code (Anthropic) to help write and debug the experiment code and to edit and polish the paper's text. 

\bibliography{custom}

\appendix

\section{Metrics and Results on Devtest}
\label{app:results}

Table~\ref{tab:metrics} defines the metrics of Section~\ref{sec:eval-metrics}. Table~\ref{tab:main} gives the results of Section~\ref{sec:res-main} on the full devtest split, and Table~\ref{tab:groups} splits hit and hit-and-correct by ask-$k$ group.

\begin{table}[t]
  \centering
  \small
  \renewcommand{\arraystretch}{1.1}
  \resizebox{\linewidth}{!}{
  \begin{tabular}{@{}p{0.17\linewidth}p{0.74\linewidth}@{}}
    \toprule
    \textbf{Metric} & \textbf{What it measures} \\
    \midrule
    \multicolumn{2}{@{}l}{\textit{Accuracy and cost}} \\
    Reward & Mean episode return (Eq.~\ref{eq:reward}), weighing correct answers against the questions spent. \\
    Accuracy & Fraction of episodes that end on the correct referent. \\
    Questions & Average number of questions asked per episode. \\
    \midrule
    \multicolumn{2}{@{}l}{\textit{When to ask}} \\
    Ask F1 & Balances precision (share of episodes with a question that were ambiguous) and recall (share of ambiguous episodes that received a question). \\

    Over-asking & Fraction of unambiguous episodes that still received a question. \\
    Under-asking & Fraction of ambiguous episodes answered without asking ($1 - {}$recall). \\
    Hallucinated certainty & Fraction of ambiguous episodes answered without asking, and wrongly. \\
    \midrule
    \multicolumn{2}{@{}l}{\textit{When to stop}} \\
    Hit & Answering after exactly as many questions as the oracle asks.\\ %
    Hit-and-correct & A hit that also ends on the correct referent. \\
    \bottomrule
  \end{tabular}}
  \caption{Evaluation metrics, as fractions of all episodes unless stated. Hit and hit-and-correct are also reported within each ask-$k$ group. Questions refused past the budget are not counted.}
  \label{tab:metrics}
\end{table}

\begin{table}[t]
\centering
\footnotesize
\setlength{\tabcolsep}{2pt}
\resizebox{\columnwidth}{!}{%
\begin{tabular}{@{}lccccc@{}}
\toprule
 & \textbf{Natural} & \textbf{SFT} & \textbf{GRPO} & \textbf{ArCHer} & \textbf{BACE} \\
\midrule
\multicolumn{6}{@{}l}{\textit{Devtest, 2,483 episodes}} \\
Ask F1 & .218 & 1.000 & 1.000\sd{.000} & 1.000\sd{.000} & 1.000\sd{.001} \\
Over & .519 & .000 & .000\sd{.000} & .000\sd{.000} & .000\sd{.000} \\
Under & .739 & .000 & .000\sd{.000} & .000\sd{.000} & .001\sd{.001} \\
Halluc. & .520 & .000 & .000\sd{.000} & .000\sd{.000} & .000\sd{.000} \\
Hit & .369 & .981 & .964\sd{.027} & .958\sd{.013} & .981\sd{.000} \\
Acc. & .782 & .990 & .983\sd{.014} & .979\sd{.006} & .992\sd{.000} \\
Qs & .79 & .43 & .39\sd{.03} & .40\sd{.02} & .43\sd{.00} \\
Reward & .486 & .938 & .926\sd{.025} & .918\sd{.011} & .940\sd{.000} \\
\bottomrule
\end{tabular}
}
\caption{End of one pass over the train split (BACE: 150 warm-start episodes, then 552 RL iterations of 20). Natural: Natural Behavior. SFT is the checkpoint every RL run starts from. Metrics as in Table~\ref{tab:metrics}. RL methods: mean over five seeds, SD in parentheses; their training episodes and optimizer steps are in Table~\ref{tab:cost}.}
\label{tab:main}
\end{table}

\begin{table}[t]
\centering
\footnotesize
\setlength{\tabcolsep}{2pt}
\begin{tabular}{@{}lccccc@{}}
\toprule
 & \textbf{Natural} & \textbf{SFT} & \textbf{GRPO} & \textbf{ArCHer} & \textbf{BACE} \\
\midrule
\multicolumn{6}{@{}l}{\textit{Ask-1, 550 episodes}} \\
Hit & .155 & .978 & .992\sd{.005} & .985\sd{.011} & .978\sd{.000} \\
H+C & .044 & .976 & .990\sd{.003} & .983\sd{.010} & .978\sd{.000} \\
\midrule
\multicolumn{6}{@{}l}{\textit{Ask-2, 184 episodes}} \\
Hit & .049 & .891 & .702\sd{.342} & .647\sd{.144} & .853\sd{.000} \\
H+C & .016 & .859 & .660\sd{.324} & .623\sd{.148} & .853\sd{.000} \\
\midrule
\multicolumn{6}{@{}l}{\textit{Ask-3, 48 episodes}} \\
Hit & .062 & .667 & .350\sd{.121} & .342\sd{.157} & .833\sd{.000} \\
H+C & .021 & .479 & .250\sd{.074} & .200\sd{.111} & .562\sd{.000} \\
\bottomrule
\end{tabular}
\caption{Hit and hit-and-correct (H+C) on {\tt devtest} by the number of questions the oracle needs, at the points of Table~\ref{tab:main}. Ask-3 includes episodes the oracle can't resolve within the budget. Columns as in Table~\ref{tab:main}. Mean over seeds, SD in parentheses.}
\label{tab:groups}
\end{table}

\section{Results on Dev}

\label{app:dev}

\begin{figure*}[!t]
  \centering
  \includegraphics[width=0.19\textwidth]{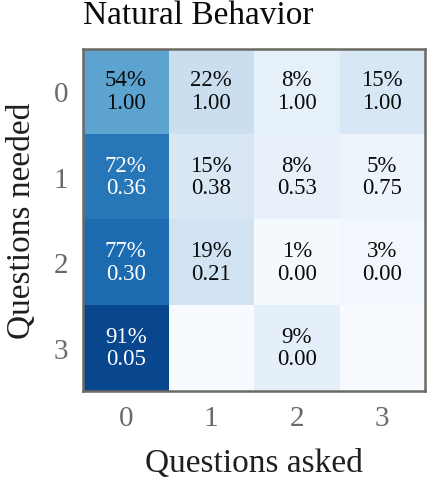}\hfill
  \includegraphics[width=0.19\textwidth]{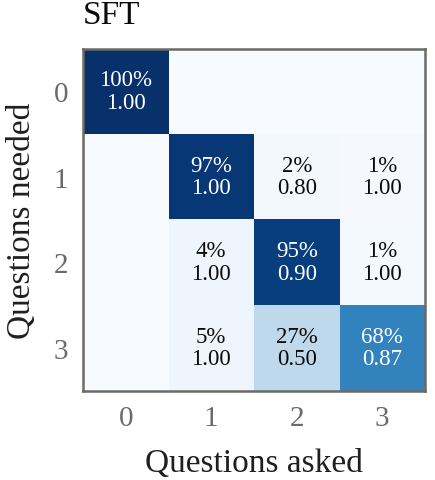}\hfill
  \includegraphics[width=0.19\textwidth]{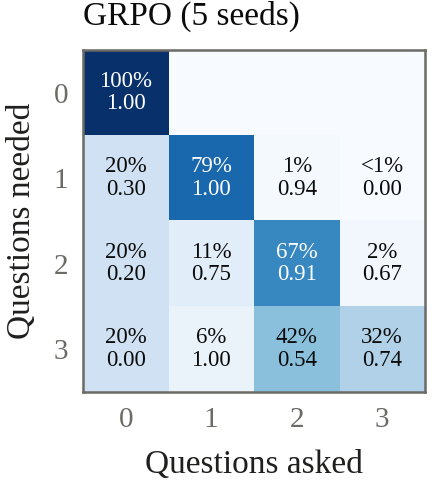}\hfill
  \includegraphics[width=0.19\textwidth]{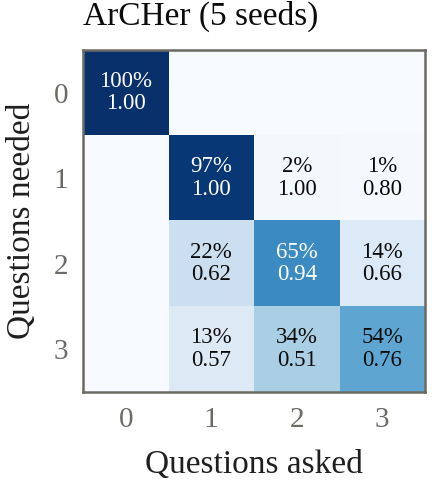}\hfill
  \includegraphics[width=0.19\textwidth]{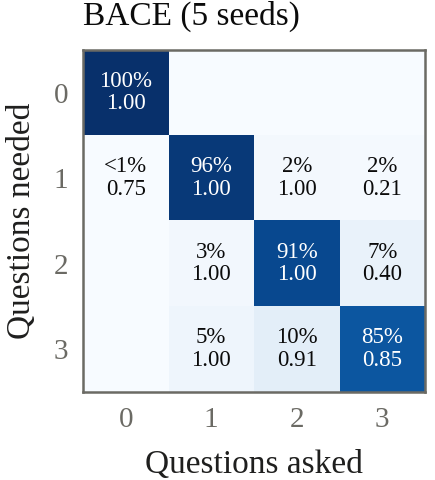}
  \caption{Figure~\ref{fig:depth-asks} on dev.}
  \label{fig:depth-asks-dev}
\end{figure*}

\begin{figure*}[!t]
  \centering
  \includegraphics[width=0.32\textwidth]{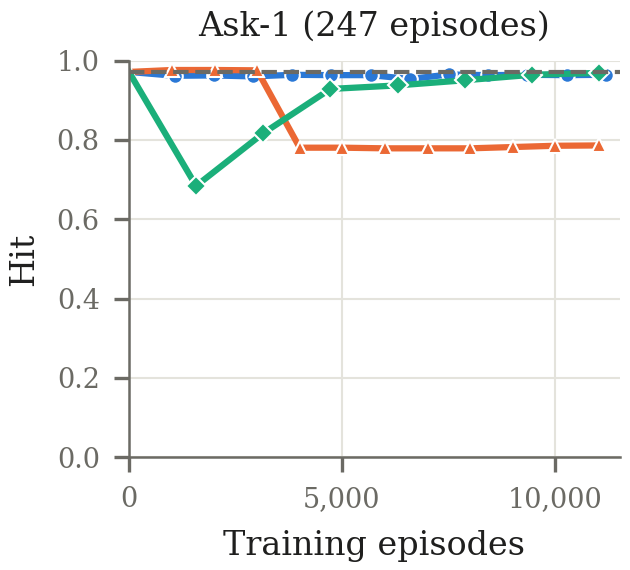}\hfill
  \includegraphics[width=0.32\textwidth]{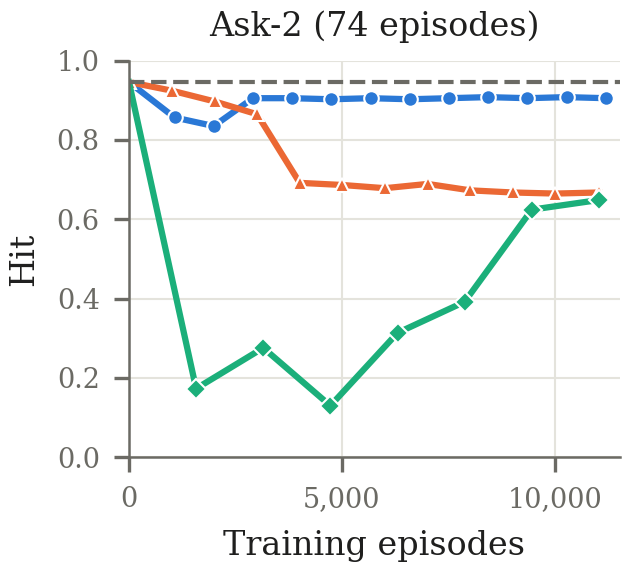}\hfill
  \includegraphics[width=0.32\textwidth]{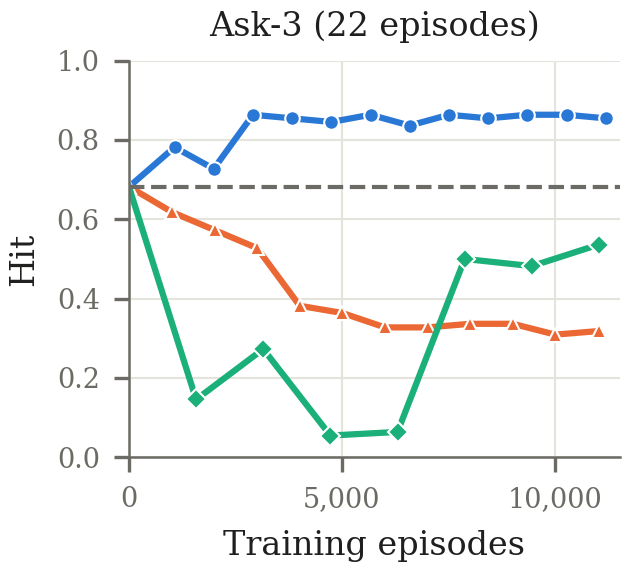}\\[2pt]
  \includegraphics[width=0.8\textwidth]{plots/curves_legend.png}
  \caption{Hit on dev against training episodes, by the number of questions the oracle needs: the share of episodes where the policy answers after exactly as many questions as the oracle. Ask-3 includes episodes the oracle can't resolve within the budget. Each line is a method's five-seed mean, starting at SFT (dashed).}
  \label{fig:curves-hit-dev}
\end{figure*}

\begin{figure*}[!t]
  \centering
  \includegraphics[width=0.32\textwidth]{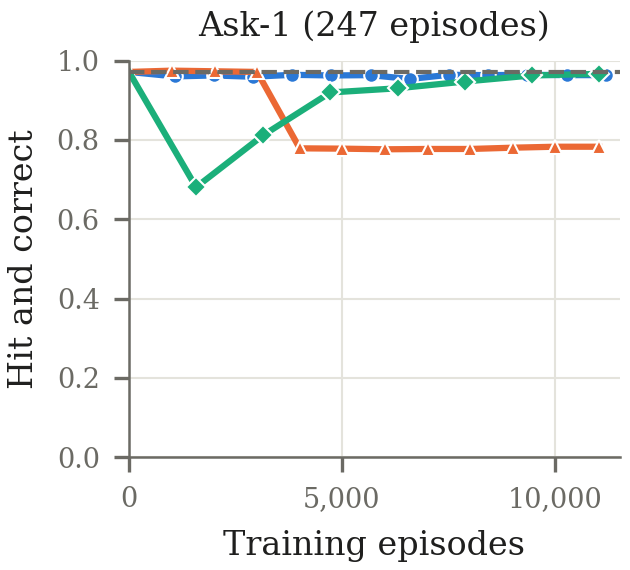}\hfill
  \includegraphics[width=0.32\textwidth]{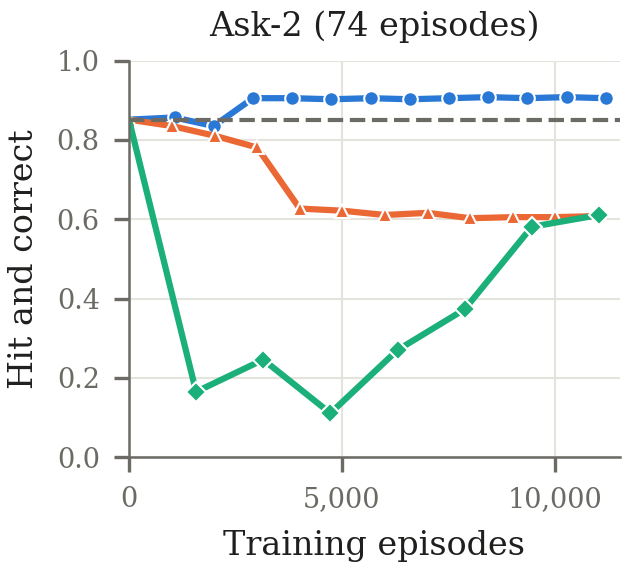}\hfill
  \includegraphics[width=0.32\textwidth]{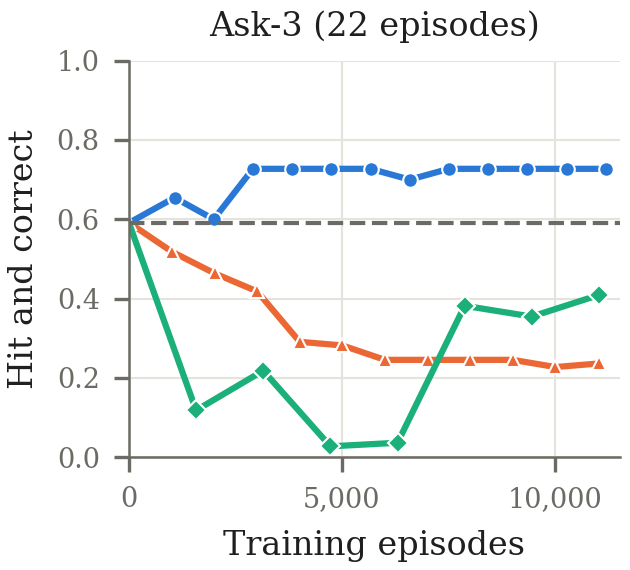}\\[2pt]
  \includegraphics[width=0.8\textwidth]{plots/curves_legend.png}
  \caption{Hit-and-correct on dev against training episodes, by the number of questions the oracle needs: a hit that also ends on the correct referent. Plots and lines as in Figure~\ref{fig:curves-hit-dev}.}
  \label{fig:curves-hit-correct-dev}
\end{figure*}

Tables~\ref{tab:main-dev} and~\ref{tab:groups-dev} and Figure~\ref{fig:depth-asks-dev} repeat Tables~\ref{tab:main} and~\ref{tab:groups} and Figure~\ref{fig:depth-asks} on the 960 dev episodes, at the same points, and Figures~\ref{fig:curves-hit-dev} and~\ref{fig:curves-hit-correct-dev} follow each method through its pass on dev. On dev, Natural Behavior asks on 45.5\% of the episodes that need no question and leaves 74.6\% of the ambiguous ones without one (Ask F1 0.245), BACE ends at 0.931 $\pm$ 0.001 and its warm start alone gives 0.880 $\pm$ 0.060, and BACE's five seeds agree on 957 of the 960 episodes.

\begin{table}[t]
\centering
\footnotesize
\setlength{\tabcolsep}{2pt}
\resizebox{\columnwidth}{!}{%
\begin{tabular}{@{}lccccc@{}}
\toprule
 & \textbf{Natural} & \textbf{SFT} & \textbf{GRPO} & \textbf{ArCHer} & \textbf{BACE} \\
\midrule
\multicolumn{6}{@{}l}{\textit{Dev, 960 episodes}} \\
Ask F1 & .245 & 1.000 & .800\sd{.447} & 1.000\sd{.000} & .999\sd{.001} \\
Over & .455 & .000 & .000\sd{.000} & .000\sd{.000} & .000\sd{.000} \\
Under & .746 & .000 & .200\sd{.447} & .000\sd{.000} & .002\sd{.002} \\
Halluc. & .507 & .000 & .148\sd{.330} & .000\sd{.000} & .001\sd{.001} \\
Hit & .390 & .979 & .904\sd{.146} & .954\sd{.015} & .980\sd{.000} \\
Acc. & .766 & .986 & .932\sd{.110} & .978\sd{.007} & .990\sd{.001} \\
Qs & .69 & .48 & .37\sd{.21} & .47\sd{.02} & .49\sd{.00} \\
Reward & .463 & .925 & .828\sd{.199} & .910\sd{.012} & .931\sd{.001} \\
\bottomrule
\end{tabular}
}
\caption{Table~\ref{tab:main} on {\tt dev}, at the same points. Columns and metrics as in Table~\ref{tab:main}. RL methods: mean over five seeds, SD in parentheses.}
\label{tab:main-dev}
\end{table}

\begin{table}[t]
\centering
\footnotesize
\setlength{\tabcolsep}{2pt}
\begin{tabular}{@{}lccccc@{}}
\toprule
 & \textbf{Natural} & \textbf{SFT} & \textbf{GRPO} & \textbf{ArCHer} & \textbf{BACE} \\
\midrule
\multicolumn{6}{@{}l}{\textit{Ask-1, 247 episodes}} \\
Hit & .150 & .972 & .786\sd{.440} & .969\sd{.009} & .964\sd{.002} \\
H+C & .057 & .972 & .783\sd{.438} & .967\sd{.009} & .964\sd{.000} \\
\midrule
\multicolumn{6}{@{}l}{\textit{Ask-2, 74 episodes}} \\
Hit & .014 & .946 & .668\sd{.376} & .649\sd{.147} & .905\sd{.000} \\
H+C & .000 & .851 & .608\sd{.341} & .611\sd{.138} & .905\sd{.000} \\
\midrule
\multicolumn{6}{@{}l}{\textit{Ask-3, 22 episodes}} \\
Hit & .000 & .682 & .318\sd{.275} & .536\sd{.235} & .855\sd{.020} \\
H+C & .000 & .591 & .236\sd{.219} & .409\sd{.185} & .727\sd{.000} \\
\bottomrule
\end{tabular}
\caption{Table~\ref{tab:groups} on {\tt dev}, at the points of Table~\ref{tab:main-dev}. Columns as in Table~\ref{tab:main}. Mean over seeds, SD in parentheses.}
\label{tab:groups-dev}
\end{table}

\section{Prompted Policies}
\label{app:prompted}

Table~\ref{tab:prompted} gives the six policies that use the instruct model without training: Natural Behavior (Section~\ref{sec:natural}) and five prompted policies. Each decodes greedily and is a single run. Direct Answer is told to answer immediately. Ask Once Then Answer asks on the first turn and answers on the second, whatever the reply. Ask Until Resolved always asks first, then keeps asking while its confidence in one candidate is below 0.6 and questions remain. Uncertainty-gated asks only when that confidence is below 0.6, so it may never ask. Confidence is a softmax over the candidates of the length-normalized log-likelihood of \texttt{object\_id: <id>}. ProCoT \citep{deng2023prompting} reasons about the ambiguity before committing to ask or answer. The prompts are in Appendix~\ref{app:prompts}.

Asking once on every episode raises accuracy over Direct Answer by 0.12, and the 617 questions it spends on unambiguous dev episodes cost it 0.06 of reward. Uncertainty-gated has the best Ask F1 and never asks when no question is needed, but it skips 39\% of the ambiguous episodes. None of the six gets above 0.79 accuracy on ask-1 episodes, 0.40 on ask-2 or 0.41 on ask-3, where SFT reaches 1.000, 0.932 and 0.636 on dev and 0.996, 0.951 and 0.729 on devtest. ProCoT's row moved by about 3 points between two runs with nothing in its inputs changed, so we treat it as less certain than the others.

\begin{table}[t]
\centering
\small
\setlength{\tabcolsep}{3pt}
\resizebox{\columnwidth}{!}{%
\begin{tabular}{@{}lccccccccc@{}}
\toprule
\textbf{Policy} & \textbf{Acc.} & \textbf{Qs} & \textbf{Reward} & \textbf{Ask F1} & \textbf{Over} & \textbf{Under} & \textbf{Ask-1} & \textbf{Ask-2} & \textbf{Ask-3} \\
\midrule
\multicolumn{10}{@{}l}{\textit{Dev, 960 episodes}} \\
Direct Answer & .761 & .00 & .523 & .000 & .000 & 1.000 & .397 & .176 & .136 \\
Natural Behavior & .766 & .69 & .463 & .245 & .455 & .746 & .393 & .270 & .045 \\
Ask Once Then Answer & .881 & 1.00 & .662 & .526 & 1.000 & .000 & .785 & .378 & .318 \\
Ask Until Resolved & .865 & 1.37 & .593 & .526 & 1.000 & .000 & .721 & .392 & .273 \\
Uncertainty-gated & .823 & .55 & .591 & .762 & .000 & .385 & .555 & .365 & .409 \\
ProCoT & .830 & 1.99 & .461 & .532 & .778 & .131 & .611 & .297 & .318 \\
\midrule
\multicolumn{10}{@{}l}{\textit{Devtest, 2,483 episodes}} \\
Direct Answer & .792 & .00 & .584 & .000 & .000 & 1.000 & .404 & .201 & .146 \\
Natural Behavior & .782 & .79 & .486 & .218 & .519 & .739 & .349 & .234 & .125 \\
Ask Once Then Answer & .892 & 1.00 & .684 & .479 & 1.000 & .000 & .782 & .370 & .333 \\
Ask Until Resolved & .881 & 1.29 & .633 & .479 & 1.000 & .000 & .742 & .337 & .333 \\
Uncertainty-gated & .845 & .44 & .646 & .749 & .000 & .402 & .593 & .304 & .312 \\
ProCoT & .855 & 1.91 & .518 & .471 & .786 & .165 & .640 & .337 & .146 \\
\bottomrule
\end{tabular}
}
\caption{The untrained policies. Ask-1, Ask-2, Ask-3: accuracy on the episodes that need one, two, and three or more questions. Other metrics as in Table~\ref{tab:metrics}.}
\label{tab:prompted}
\end{table}

\section{Synthetic-Episode Check}
\label{app:synthetic}

We discard a synthetic episode if its utterance no longer fits the smaller candidate set: when the utterance names a type, color or pattern that one of the kept candidates lacks, or contains a count word (``both'', ``the two''), since the number of candidates has changed. We also discard duplicates (the same template with the same candidate subset) and cap each template at 20 synthetic episodes per depth, to avoid having a few scenes dominate.

We test whether a simple classifier can tell synthetic episodes from real ones when it reads exactly what the text policy reads. Each episode is its rendered observation (Appendix~\ref{app:prompts}), lowercased and split into words, numbers and punctuation marks. The features are the unigrams and bigrams of those tokens hashed into $2^{18}$ buckets, with value $\log(1+\text{count})$ and each row scaled to unit length. A logistic regression with an L2 penalty of $10^{-5}$ is fit by 500 steps of full-batch gradient descent on a random 80\% of the episodes, and we report its AUC on the other 20\%.

The real side is the ambiguous training episodes. At each depth they are paired with as many synthetic training episodes of the same depth, drawn at random, and the pooled comparison pairs all 3,677 of them with as many synthetic ones. Depth is counted with the budget of three questions, as in the ask-$k$ groups. As a control, the real episodes of each depth are split at random into two halves with arbitrary labels, which gives the classifier's noise floor.

Table~\ref{tab:synthetic} shows the result. The pooled AUC of 0.70 mostly reflects depth: synthetic episodes are deeper by design, and deeper episodes have more candidates (a median of 7 against 4). Matched on depth, the classifier stays within 0.06 to 0.09 of its noise floor. What it does pick up is consistent with those larger candidate sets and with vaguer utterances: 35\% of synthetic utterances name a type, color or pattern against 53\% of real ones, because the large annotated sets that deep episodes need usually come from vague requests.

\begin{table}[t]
\centering
\small
\setlength{\tabcolsep}{4pt}
\begin{tabular}{@{}lcccc@{}}
\toprule
 & \textbf{Pooled} & \textbf{Depth 1} & \textbf{Depth 2} & \textbf{Depth 3} \\
\midrule
Real episodes & 3,677 & 2,613 & 841 & 223 \\
Real vs.\ real & 0.52 & 0.50 & 0.54 & 0.57 \\
Real vs.\ synthetic & 0.70 & 0.59 & 0.60 & 0.63 \\
\bottomrule
\end{tabular}
\caption{Held-out AUC of a bag-of-n-grams classifier separating two sets of training episodes (a value of 0.5 means the two are indistinguishable). Each comparison uses as many synthetic episodes as real ones. Real vs.\ real splits the real episodes at random.}
\label{tab:synthetic}
\end{table}

\section{Hyperparameters}
\label{app:hyper}

Tables~\ref{tab:hp-sft} to~\ref{tab:hp-bace} list the settings of the four trained methods. All of them add LoRA adapters to every attention and MLP projection of Qwen2.5-7B-Instruct (dropout 0.05), use AdamW with linearly decaying learning rates, and run with $\lambda = 0.1$ and a budget of three questions. Prompts are limited to 1,280 tokens and a turn to 64 new tokens. The RL methods merge the SFT adapter into the backbone before adding their own, and the critics use $\gamma = 1$.

SFT's targets follow the oracle's chain, with its attribute recomputed on the survivors after every reply, since the first reply can make a weak attribute decisive or a strong one useless; if the question budget runs out first, the target answers with the first survivor.

We implement every method with PyTorch 2.7.1, Hugging Face Transformers 5.14.1 and PEFT 0.20.0. The PPO clip and the GRPO KL coefficient follow \citet{schulman2017proximal} and \citet{shao2024deepseekmath}, and we changed a setting only when a run failed. Section~\ref{sec:archer} and Appendix~\ref{app:bace-stability} describe those changes.

In ArCHer, the actor's advantage is the smaller of the two Q values minus the smaller of the two V values, with Polyak-averaged targets. The SFT policy makes the ask/answer choice during the 10 critic-only iterations; without this warm-up, the actor fell into repeating tokens within a few iterations. We add a KL penalty to the SFT policy ($\beta = 0.04$), which the original doesn't use with its GPT-2 actor. Because the SFT-initialized actor, given \texttt{ASK:}, often wrote an object ID instead of an attribute, so that more than half of its questions revealed nothing, we take the attribute from the actor's own probabilities over the askable attributes and let it write only the question text. Each critic head layer-normalizes the backbone's hidden state; without this, almost all of a head's units stopped firing early in training and Q came out nearly the same for every state and action. 

In BACE, during RL the chooser samples between the finalists from a softmax over their values at temperature 0.5, mixed with a uniform choice with weight $\epsilon = 0.15$, so that both branches keep being tried and both Q values keep being corrected by real returns. Once the three questions are spent, the question branch is masked, and the episode ends on the chooser's own answer rather than a forced one. The target critic is a copy of the whole critic, its LoRA adapter and its four heads, replaced at the end of every RL iteration; after the warm start only the heads are copied, so during the first RL iteration the target's adapter is still the untrained one. Each V head is fit to the target critic's Q value of the finalist the chooser would pick greedily, so V estimates the value of the policy we evaluate.

\begin{table}[t]
\centering
\small
\begin{tabular}{@{}ll@{}}
\toprule
Training pool & real + synthetic, 29,635 episodes \\
Examples & 54,405 of 74,110 (within 1,280 tokens) \\
LoRA rank / $\alpha$ & 16 / 32 \\
Learning rate & $10^{-4}$ \\
Batch & 4, gradient accumulation 4 \\
Epochs & 1 (3,400 optimizer steps) \\
Question text & one template per attribute \\
\bottomrule
\end{tabular}
\caption{SFT.}
\label{tab:hp-sft}
\end{table}

\begin{table}[t]
\centering
\small
\begin{tabular}{@{}ll@{}}
\toprule
LoRA rank / $\alpha$ & 16 / 32 \\
Rollouts per episode ($K$) & 4 \\
Sampling temperature & 0.8 \\
Episodes per batch & 8 (gradient accumulation 4) \\
PPO clip $\epsilon$ / passes per batch & 0.2 / 2 \\
KL coefficient to SFT ($\beta$) & 0.04 \\
Learning rate & $5\times10^{-5}$ \\
Training episodes & 11,035 (one pass) \\
\bottomrule
\end{tabular}
\caption{GRPO (trajectory level).}
\label{tab:hp-grpo}
\end{table}

\begin{table}[t]
\centering
\small
\resizebox{\columnwidth}{!}{%
\begin{tabular}{@{}ll@{}}
\toprule
Actor LoRA rank / $\alpha$ & 16 / 32 \\
Actor learning rate & $5\times10^{-5}$ \\
Actor KL coefficient to SFT & 0.04 \\
Actor schedule & 10 critic-only iterations, then 3 steps \\
 & of 4 transitions per iteration (54) \\
Actor sampling temperature & 0.8 \\
Question attribute & sampled from the actor (argmax at evaluation) \\
Token-level baseline & MLP (256), learning rate $10^{-3}$ \\
Critic LoRA rank / $\alpha$ & 8 / 16 \\
Critic heads & 2 Q/V pairs, layer norm, MLP (512) \\
Critic learning rates & adapter $10^{-4}$, heads $10^{-3}$ \\
Target critic & Polyak, $\tau = 0.005$ \\
Iterations & 28 of 394 episodes (11,032) \\
Critic transitions per iteration & 800, in steps of 4 \\
Replay buffer & 5,000 transitions \\
Exploration & softmax over Q, $T = 0.5$, $\epsilon = 0.15$ \\
Gradient clipping & actor 1.0, critic 10 \\
\bottomrule
\end{tabular}
}
\caption{ArCHer.}
\label{tab:hp-archer}
\end{table}

\begin{table}[t]
\centering
\small
\resizebox{\columnwidth}{!}{%
\begin{tabular}{@{}ll@{}}
\toprule
Generator & SFT checkpoint, frozen \\
Prompt & ArCHer's, plus the candidates in $B$ \\
Samples per branch & 2, temperature 0.8 \\
Finalist & random in training, highest length- \\
 & normalized log-probability at evaluation \\
Critic LoRA rank / $\alpha$ & 8 / 16 \\
Critic heads & 2 Q/V pairs; hidden state and features \\
 & normalized apart, features to 64; MLP (512) \\
Learning rates & adapter $10^{-4}$, heads $10^{-3}$, \\
 & decaying to 0.775 of the start \\
Q target & Equation~\ref{eq:bace-q} \\
V target & target Q of the greedy finalist \\
Target critic & full copy after every iteration (50 steps) \\
Exploration & softmax over Q, $T = 0.5$, $\epsilon = 0.15$ \\
Warm start & 150 episodes, 2 playouts per branch \\
 & on paired seeds, half of the states \\
 & with $|B| > 1$, 6 epochs (Eq.~\ref{eq:bace-warm}) \\
RL iterations & 552 of 20 episodes (8 or 100 for the shorter runs) \\
Critic transitions per iteration & 200, in steps of 4 \\
Replay buffer & 5,000 transitions \\
Gradient clipping & 10 \\
Reported point & end of the pass (11,190 episodes) \\
\bottomrule
\end{tabular}
}
\caption{BACE.}
\label{tab:hp-bace}
\end{table}

\section{Training Cost}
\label{app:cost}

Table~\ref{tab:cost} counts what each method uses to train. Environment steps are the actions the policy takes in the environment while training, both questions and answers. For BACE they include the warm start's playouts: at each warm-start state the environment is cloned four times, two answer playouts of one step and two question playouts of two steps (the question and a greedy answer), six steps in all, and the generator samples 12 completions. For GRPO it is about 58,000 on every seed. GPU hours are the wall time of the training run on one NVIDIA RTX PRO 6000 (96\,GB), without evaluation. Over the five seeds, the reported training runs took 135 to 142 GPU hours: 3.3 for SFT, 15 to 18 for GRPO, 26 for ArCHer and 91 to 95 for BACE. Counting the warm start's playouts, BACE's pass takes about 22,400 environment steps, against about 18,400 for ArCHer. ArCHer is still improving at the end of its pass, while BACE reaches its final level after a quarter of it.

\begin{table}[t]
\centering
\small
\setlength{\tabcolsep}{3pt}
\resizebox{\columnwidth}{!}{%
\begin{tabular}{@{}llccccc@{}}
\toprule
\textbf{Method} & \textbf{Learns from} & \textbf{Episodes} & \textbf{Env.\ steps} & \textbf{Opt.\ steps} & \textbf{Params} & \textbf{GPU h} \\
\midrule
SFT & oracle labels & 29,635 & -- & 3,400 & 40.4M & 3.3 \\
GRPO & reward & 11,035 & 58.1k--58.7k & 38--66 & 40.4M & 3.4--3.5 \\
ArCHer & reward & 11,032 & 18.3k--18.5k & 5,654 & 67.9M & 5.2 \\
BACE & reward & 11,190 & 22.3k--22.5k & 28,164--28,308 & 27.7M & 18.2--19.0 \\
\bottomrule
\end{tabular}
}
\caption{Training cost, as a range over seeds where it varies. Opt.\ steps: optimizer steps on every trained parameter, BACE's 564 to 708 warm-start steps included. Params: trained parameters (ArCHer: actor 40.4M, critic adapter 20.2M, heads 7.4M; BACE: critic adapter 20.2M, heads 7.5M). The RL methods start from the SFT checkpoint, whose cost comes on top of theirs. BACE's environment steps are counted on seeds 0, 1 and 7.}
\label{tab:cost}
\end{table}

\section{Prompts}
\label{app:prompts}

Every policy sends a single user message: an instruction that depends on the policy, a blank line, and the observation. After the first question the observation ends with a \texttt{Clarification so far:} block that lists each question with the scripted user's reply. Figure~\ref{fig:prompt-obs} shows a real dev observation and the replies the scripted user gives, and Figure~\ref{fig:prompt-instr} the instructions. The attribute list at the end of the ask instructions is shared by every policy that writes its question as \texttt{<attribute> | <question>}. Its format example names \texttt{color}, which the text environment doesn't offer; it is the same in every run we report.

SFT is trained to write one line per state, \texttt{ANSWER: <id>} or \texttt{ASK: <attribute> | <question>}, where the question comes from a fixed template for each attribute, such as \texttt{brand | What brand is it?} or \texttt{left\_right | Is it on the left or the right?}. GRPO uses SFT's instruction. In ArCHer the critic picks \texttt{ASK:} or \texttt{ANSWER:}, the chosen prefix is appended to the prompt, the attribute is taken from the actor's probabilities over the askable attributes, and the actor writes the rest of the line. BACE's generator and chooser read ArCHer's instruction with the prefix forced once for each branch, and its observation ends with one more line that lists the candidates still in $B$, such as \texttt{Still possible (2): [46, 49]}.

Natural Behavior gets no format, so we classify its reply (at most 64 new tokens) afterwards. A reply is a question if it ends with a question mark or starts with a question word (is, are, do, does, did, can, could, would, which, what, who, where, when, why, how). A question is mapped to an attribute by the first keyword group it contains, checked in order: available sizes (``available size'', ``sizes available'', ``what sizes'', ``in stock''), size, brand (``brand'', ``make''), price (``price'', ``cost'', ``expensive'', ``cheap'', ``how much''), customer review (``review'', ``rating'', ``rated''), the visual attributes, left/right (``left'', ``right'') and up/down (``up'', ``down'', ``top'', ``bottom'', ``higher'', ``lower''). A question that matches nothing, or whose first match is a visual attribute, reveals nothing and still costs $\lambda$. Any other reply is an answer, and its referent is the first number in the reply that is a candidate id, or the first listed candidate if there is none. Two dev examples:
\begin{itemize}
  \item ``The jeans you're asking about are priced at \$139.99. Would you like to add them to your cart or try another pair?'' (an episode that needs two questions). It ends with a question mark, and ``price'' maps it to a price question, so the scripted user replies with the referent's price. The reply had already assumed which jeans the user meant.
  \item ``The shirt hanging on the right corresponds to object 5, which has a size of XL.'' (an episode that needs one question). This is an answer, with object 5 as the referent.
\end{itemize}
The referent rule also reads numbers inside prices: ``The brown hoodies on either side of the table are priced at \$179.99 each. Would you like to see them? They are both XL size and have a rating of 4.0.'' is an answer that names object 99, because 99 is a candidate id. On dev this happens in 48 of Natural Behavior's 927 answers.

\begin{figure*}[t]
\scriptsize
\begin{verbatim}
Dialogue history:
User: Please show me some blouses.
Assistant: What do you think of the white and black blouse, the black blouse, or the light grey blouse?
User: I want a tank top that will go best with my wardrobe. Which of these will do the trick?
Assistant: You may take a look at the yellow and white tank top that matches your criteria.

User: How highly rated is that one?

Candidate objects (7):
- object 46: availableSizes=['M'], brand=Brain Puzzles, customerReview=2.9, position=(317, 534), price=39.99, size=M
- object 47: availableSizes=['S', 'XS', 'M'], brand=Art News Today, customerReview=4.7, position=(382, 585), price=59.99,
    size=S
- object 48: availableSizes=['XXL', 'XS', 'L', 'M', 'S', 'XL'], brand=Cats Are Great, customerReview=4.4,
    position=(656, 546), price=4.99, size=XS
- object 49: availableSizes=['M'], brand=Brain Puzzles, customerReview=2.9, position=(808, 566), price=39.99, size=M
- object 50: availableSizes=['S'], brand=Downtown Consignment, customerReview=3.7, position=(1356, 574), price=39.99,
    size=S
- object 51: availableSizes=['L', 'XL', 'M', 'XS'], brand=Downtown Consignment, customerReview=3.1, position=(862, 634),
    price=209.99, size=XL
- object 52: availableSizes=['L'], brand=River Chateau, customerReview=3.7, position=(136, 877), price=139.99, size=L
Spatial relation (left): {48: [47, 46], 47: [46], 49: [48, 47, 46], 50: [51, 49, 48], 51: [49, 48, 47]}
Spatial relation (right): {46: [47, 48, 49], 48: [49, 51, 50], 47: [48, 49, 51], 49: [51, 50], 51: [50], 52: [50]}
Spatial relation (up): {52: [46, 47]}
Spatial relation (down): {46: [52], 47: [52]}

Clarification so far:
Q: What sizes are available?
A: It's M.
\end{verbatim}
\vspace{-0.5em}
\textit{Scripted user replies.} Listed attribute: \texttt{It's <value>.} (a list for available sizes); no recorded value: \texttt{I don't have that information.}; half of the scene: \texttt{It's the one on the left.} / \texttt{right.} / \texttt{at the top.} / \texttt{at the bottom.}; neighbor: a yes or no to whether another remaining candidate sits on that side of the referent (left, right, up, down); a question that maps to no attribute: \texttt{Sorry, I didn't understand the question. Could you rephrase it?}
\caption{A dev observation (the referent is object 49, and the oracle's first question is about available sizes), with the clarification block as it reads after one question. Long lines are wrapped here. Only the attributes a text policy may see are listed; the position is the centre of the object's bounding box.}
\label{fig:prompt-obs}
\end{figure*}

\begin{figure*}[t]
\scriptsize
\textit{Natural Behavior.}
\begin{verbatim}
You are a shopping assistant helping a customer. Here is the conversation so far and the candidate objects you can see.
Respond to the user.
\end{verbatim}
\textit{Direct Answer.}
\begin{verbatim}
You are resolving which object a user is referring to. Given the candidate objects below, answer immediately with the single
object id you believe is correct. Do not ask any clarifying questions. Respond with exactly: "object_id: <id>".
\end{verbatim}
\textit{Ask Once Then Answer, Ask Until Resolved and Uncertainty-gated: question turn, answer turn, and the confidence call.}
\begin{verbatim}
You are resolving which object a user is referring to among several candidates. Ask ONE clarifying question that would best
help you figure out which candidate the user means. Respond with only one line.

<attribute list>

Given the candidate objects and the clarification exchange above, answer with the single object id you believe is correct:
"object_id: <id>".

You are resolving which object a user is referring to. Read the candidate objects below and judge which one the user means.
\end{verbatim}
\textit{SFT and GRPO. ProCoT replaces the opening paragraph, before the two options, with the second block.}
\begin{verbatim}
You are resolving which object a user is referring to among several candidates. Decide whether the request is ambiguous
among the candidates below and, if so, whether asking would help. Respond with exactly one line and nothing else:
- "ANSWER: <object id>" if you can resolve it (or must just guess) -- the id number alone.
- "ASK: <attribute> | <clarification question>" if it is genuinely ambiguous and asking would help.

<attribute list>

You are resolving which object a user is referring to among several candidates. Think step by step about whether the user's
request is ambiguous among the candidate objects below, and whether that ambiguity can be resolved from the dialogue history
and candidate attributes alone. Then commit to exactly one of the following as your final line:
\end{verbatim}
\textit{ArCHer, and BACE's generator and chooser.}
\begin{verbatim}
You are a shopping assistant helping a customer identify exactly one object among the candidates listed below.

Your reply has already been started for you, with either "ASK: " or "ANSWER: ". Continue that line and stop. Do not repeat
the prefix, do not write the other one, and do not add any explanation.

Continuing "ASK: " -- <attribute list>
Never resolve the customer's request in place of asking.

Continuing "ANSWER: " -- write only the id number of the single candidate the customer means, and nothing else.
Example: 12
\end{verbatim}
\textit{The attribute list.}
\begin{verbatim}
Write the clarifying question in exactly this form:
    <attribute> | <question>
<attribute> must be copied character-for-character from this list:
  availableSizes, brand, customerReview, price, size
      -- properties of the object itself
  left_right, up_down
      -- which half of the scene the object sits in
  left, right, up, down
      -- whether another candidate sits on that side of it
Choose the attribute whose values differ most among the candidates. <question> must be one natural question to the customer
about that attribute. Never invent an attribute name, and never name an object id or number inside the question.
Format example: color | What color is the jacket you mean?
\end{verbatim}
\caption{The instructions that precede the observation. Long lines are wrapped here.}
\label{fig:prompt-instr}
\end{figure*}

\section{SFT by Training Pool}
\label{app:sft-pool}

Table~\ref{tab:sft-pool} trains the same SFT recipe on three pools: a fixed sample of 150 real episodes (the pilot pool), the 11,035 real training episodes with a resolvable chain, and the real episodes with the synthetic ones added. The decision to ask is learned from very little data: 150 episodes already give an Ask F1 of 0.978. What more oracle data improves is everything after that decision, and multi-question accuracy most of all, which rises from 0.354 to 0.708 with the real episodes and to 0.875 only with the synthetic ones. The checkpoint trained on all of them is the starting point of every RL method and BACE's generator.

\begin{table}[t]
\centering
\small
\setlength{\tabcolsep}{3pt}
\resizebox{\columnwidth}{!}{%
\begin{tabular}{@{}lcccccc@{}}
\toprule
\textbf{Pool} & \textbf{Episodes} & \textbf{Ask F1} & \textbf{Acc.} & \textbf{Qs} & \textbf{Reward} & \textbf{Multi} \\
\midrule
\multicolumn{7}{@{}l}{\textit{Dev}} \\
Pilot & 150 & .978 & .857 & .54 & .661 & .354 \\
Real & 11,035 & 1.000 & .965 & .47 & .883 & .708 \\
Real + synth. & 29,635 & 1.000 & .986 & .48 & .925 & .865 \\
\midrule
\multicolumn{7}{@{}l}{\textit{Devtest}} \\
Pilot & 150 & .973 & .867 & .46 & .688 & .330 \\
Real & 11,035 & 1.000 & .974 & .41 & .908 & .710 \\
Real + synth. & 29,635 & 1.000 & .990 & .43 & .938 & .880 \\
\bottomrule
\end{tabular}
}
\caption{SFT trained on each pool. Multi: accuracy on a fixed sample of up to 100 episodes that need two or more questions (all 96 on dev, a sample of 100 on devtest).}
\label{tab:sft-pool}
\end{table}

\section{BACE Ablations}
\label{app:bace-ablations}

Tables~\ref{tab:bace-line} and~\ref{tab:bace-gen} change one part of BACE at a time, on seeds 0, 1 and 7, with 100 RL iterations (2,000 episodes) instead of the full pass. Without the \texttt{Still possible} line, which the generator and the chooser read in the same prompt, every seed ends a little lower. It asks on the same episodes and as often, and loses accuracy on episodes that need two or more questions; seed 0 also stops early on most ask-3 episodes and hardly asks at iteration 25. It stays ahead of GRPO and ArCHer on ask-2 and ask-3 episodes (Table~\ref{tab:groups}) but below SFT, so BACE keeps the line. In the trained chooser of the full pass, the decision comes from $\varphi$ and $\psi$ (Table~\ref{tab:bace-routes}): changing the line leaves it as it is, and computing the features as if no candidate had been ruled out makes the chooser ask again where $B$ is down to one candidate. The line speeds up training and adds accuracy, and when to ask comes from the belief the chooser tracks.

With the SFT checkpoint trained on real episodes only as the generator, which is weaker on its own (Table~\ref{tab:sft-pool}), BACE ends above the control and above the SFT checkpoint of the main comparison without any synthetic episode. Its questions are weaker, so it more often answers late, but it keeps asking until one candidate is left and is the most accurate of the three.

\begin{table}[t]
\centering
\small
\setlength{\tabcolsep}{3pt}
\resizebox{\columnwidth}{!}{%
\begin{tabular}{@{}lcccccc@{}}
\toprule
 & \multicolumn{2}{c}{\textbf{Reward}} & \multicolumn{4}{c}{\textbf{Devtest}} \\
\cmidrule(lr){2-3} \cmidrule(lr){4-7}
 & \textbf{Dev} & \textbf{Devtest} & \textbf{Hit} & \textbf{Acc.} & \textbf{Ask-2 H+C} & \textbf{Ask-3 H+C} \\
\midrule
SFT (real + synth.) & .925 & .938 & .981 & .990 & .859 & .479 \\
BACE, with the line & .930\sd{.004} & .935\sd{.009} & .977\sd{.007} & .989\sd{.005} & .832\sd{.038} & .514\sd{.084} \\
BACE, without the line & .917\sd{.005} & .922\sd{.005} & .974\sd{.007} & .982\sd{.003} & .804\sd{.028} & .361\sd{.241} \\
\bottomrule
\end{tabular}
}
\caption{BACE with and without the \texttt{Still possible} line in the prompt that the generator and the chooser share, after 100 RL iterations (2,000 episodes); mean over seeds 0, 1 and 7, SD in parentheses. SFT (real + synth.) is the generator of both rows and of the main comparison.}
\label{tab:bace-line}
\end{table}

\begin{table}[t]
\centering
\small
\begin{tabular}{@{}lccc@{}}
\toprule
 & \multicolumn{2}{c}{\textbf{Flipped}} & \\
\cmidrule(lr){2-3}
\textbf{Change} & \textbf{All} & \textbf{$b = 1$} & \textbf{Gap} \\
\midrule
No line & 0 & 0 & .0003 \\
Line lists every candidate & 0 & 0 & .0002 \\
$\varphi, \psi$ from all candidates & 220 & 218 & .2988 \\
\bottomrule
\end{tabular}
\caption{The trained chooser of seed 0's full pass on 300 dev episodes, at the 690 states where it may ask (272 with $b = 1$), with one route of the belief changed and the sampled candidates fixed. Flipped: decisions that change. Gap: mean absolute change in the ask-minus-answer value gap.}
\label{tab:bace-routes}
\end{table}

\begin{table}[t]
\centering
\small
\setlength{\tabcolsep}{3pt}
\resizebox{\columnwidth}{!}{%
\begin{tabular}{@{}lcccccc@{}}
\toprule
 & \multicolumn{2}{c}{\textbf{Reward}} & \multicolumn{4}{c}{\textbf{Devtest}} \\
\cmidrule(lr){2-3} \cmidrule(lr){4-7}
 & \textbf{Dev} & \textbf{Devtest} & \textbf{Hit} & \textbf{Acc.} & \textbf{Ask-2 H+C} & \textbf{Ask-3 H+C} \\
\midrule
SFT (real + synth.) & .925 & .938 & .981 & .990 & .859 & .479 \\
BACE, generator SFT (real + synth.) & .930\sd{.004} & .935\sd{.009} & .977\sd{.007} & .989\sd{.005} & .832\sd{.038} & .514\sd{.084} \\
BACE, generator SFT (real) & .936\sd{.000} & .945\sd{.000} & .954\sd{.000} & .995\sd{.000} & .739\sd{.000} & .604\sd{.000} \\
\bottomrule
\end{tabular}
}
\caption{BACE with each SFT checkpoint as its generator, after 100 RL iterations (2,000 episodes); mean over seeds 0, 1 and 7, SD in parentheses. SFT (real): trained on real episodes only, scored on its own in Table~\ref{tab:sft-pool}. The first BACE row is the run with the line in Table~\ref{tab:bace-line}.}
\label{tab:bace-gen}
\end{table}

\section{BACE per Seed}
\label{app:bace-seeds}

Besides the full pass, we trained the same recipe for 8 RL iterations (310 episodes) on each seed, which we call the 8-iteration runs. Table~\ref{tab:bace-seeds} gives each seed's reward after the warm start, after RL iterations 5 and 8, and at the end of the full pass. Seeds 0 and 1 of the 8-iteration runs are reruns with the features of Equation~\ref{eq:bace-phi}; their first runs also gave the chooser a flag for whether $B$ was known to equal the survivor set, which is set on almost every state and which we dropped. The rerun of seed 0 is the one run that ends badly: from iteration 3 on it swings between asking and not asking (Appendix~\ref{app:bace-stability}), and it finishes at 0.836 with under-asking of 0.216.

Several checkpoints score exactly 0.932 on dev: seed 1 at iteration 5, seed 123 at the end, and the full passes of seeds 0, 1 and 7. With the generator frozen and evaluation seeded per episode, two choosers that make the same decisions produce the same dialogues, and these all ask while more than one candidate is left. On devtest the three full passes score 0.940 and the other two 0.939.

\begin{table}[t]
\centering
\small
\setlength{\tabcolsep}{3pt}
\resizebox{\columnwidth}{!}{%
\begin{tabular}{@{}lccccccccc@{}}
\toprule
 & \multicolumn{5}{c}{\textbf{8-iteration runs}} & \multicolumn{4}{c}{\textbf{Full pass}} \\
\cmidrule(lr){2-6} \cmidrule(lr){7-10}
 & \textbf{Warm} & \multicolumn{2}{c}{\textbf{It.\ 5}} & \multicolumn{2}{c}{\textbf{End (it.\ 8)}} & \multicolumn{2}{c}{\textbf{Warm}} & \multicolumn{2}{c}{\textbf{End}} \\
\cmidrule(lr){2-2} \cmidrule(lr){3-4} \cmidrule(lr){5-6} \cmidrule(lr){7-8} \cmidrule(lr){9-10}
\textbf{Seed} & \textbf{dev} & \textbf{dev} & \textbf{devtest} & \textbf{dev} & \textbf{devtest} & \textbf{dev} & \textbf{devtest} & \textbf{dev} & \textbf{devtest} \\
\midrule
0 & .923 & .905 & .900 & .836 & .827 & .923 & .932 & .932 & .940 \\
1 & .932 & .932 & .939 & .909 & .918 & .775 & .772 & .932 & .940 \\
7 & .931 & .931 & .940 & .930 & .939 & .910 & .912 & .932 & .940 \\
17 & .932 & .922 & .931 & .927 & .935 & .886 & .876 & .930 & .940 \\
123 & .930 & .930 & .938 & .932 & .939 & .905 & .913 & .930 & .941 \\
\midrule
Mean & .929 & .924 & .930 & .907 & .912 & .880 & .881 & .931 & .940 \\
SD & .004 & .011 & .017 & .041 & .048 & .060 & .064 & .001 & .000 \\
\bottomrule
\end{tabular}
}
\caption{BACE's reward per seed. Warm: the chooser after its warm start, before RL; the 8-iteration runs fit their own warm start.}
\label{tab:bace-seeds}
\end{table}

\section{BACE Stability}
\label{app:bace-stability}

Table~\ref{tab:bace-gap} follows the chooser through the 8 RL iterations on two seeds, scored on dev after each iteration. The gap is the chooser's value of asking minus its value of answering at a state, each the smaller of its two Q heads, and we report its median over the states with more than one candidate left (ambiguous) and over those with one (resolved). On seed 7 the ambiguous gap stays between $+1.16$ and $+2.15$ and resolved states sit near zero or below, so the chooser asks on 98\% of ambiguous states at every iteration. (The other 2\% are states where the budget is spent and only an answer is allowed.) On the rerun of seed 0 the two kinds of state stop being separated after iteration 2. From then on the ambiguous gap stays within $0.08$ of zero, and the greedy choice on those states flips with it: the chooser asks on none of them after iterations 3 and 6 and on most of them after iteration 5. Seed 7's dip after iteration 1 has a different cause: its resolved states briefly get a positive gap, and it asks on 98.5\% of the episodes that need no question.

Before RL, every run logs two diagnostics of the warm-started chooser on the warm-start states: its greedy ask rate, and its sign agreement, the share of states where the sign of its ask-minus-answer gap matches the sign of the measured return gap $\Delta G$ (Equation~\ref{eq:bace-warm}). We used to abort a run when the ask rate fell outside $[0.25, 0.60]$ or the sign agreement below 0.60. To see whether this check predicts how a run ends, we trained the 8-iteration recipe four times on each of seeds 0 and 7 without the abort (Table~\ref{tab:bace-gate}). The only draw that failed the check ended at 0.932 dev reward, as high as any, while a draw that passed it ended at 0.712 with under-asking of 0.531. Seed 0's four draws end between 0.712 and 0.932. The check doesn't predict the outcome, so we no longer use it to stop runs. The draws differ even though they share the seed: the warm-start fit isn't reproducible from run to run because of nondeterministic GPU kernels, and two fits on the same states with the same targets give gaps that correlate between 0.64 and 0.91.

\begin{table}[t]
\centering
\small
\setlength{\tabcolsep}{3pt}
\resizebox{\columnwidth}{!}{%
\begin{tabular}{@{}lcccccccc@{}}
\toprule
 & \multicolumn{4}{c}{\textbf{Seed 0 (rerun)}} & \multicolumn{4}{c}{\textbf{Seed 7}} \\
\cmidrule(lr){2-5} \cmidrule(lr){6-9}
\textbf{It.} & \textbf{Amb.} & \textbf{Res.} & \textbf{Asks} & \textbf{Reward} & \textbf{Amb.} & \textbf{Res.} & \textbf{Asks} & \textbf{Reward} \\
\midrule
Warm & $+1.71$ & $-0.09$ & .95 & .923 & $+1.16$ & $-0.05$ & .98 & .931 \\
1 & $+1.86$ & $-0.06$ & .96 & .931 & $+1.59$ & $+0.03$ & .98 & .791 \\
2 & $+1.12$ & $-0.37$ & .95 & .926 & $+1.54$ & $-0.05$ & .98 & .932 \\
3 & $-0.08$ & $-0.15$ & .00 & .500 & $+1.57$ & $-0.14$ & .98 & .932 \\
4 & $-0.01$ & $-0.14$ & .21 & .591 & $+1.68$ & $-0.12$ & .98 & .932 \\
5 & $+0.06$ & $-0.12$ & .84 & .905 & $+1.64$ & $-0.11$ & .98 & .931 \\
6 & $-0.06$ & $-0.12$ & .00 & .500 & $+1.77$ & $-0.14$ & .98 & .932 \\
7 & $-0.05$ & $-0.21$ & .04 & .519 & $+2.15$ & $-0.12$ & .98 & .930 \\
8 & $+0.01$ & $-0.08$ & .68 & .836 & $+1.74$ & $-0.12$ & .98 & .930 \\
\bottomrule
\end{tabular}
}
\caption{The chooser on dev after each RL iteration. Amb., Res.: median ask-minus-answer value gap on states with more than one candidate left and with one. Asks: share of the ambiguous states where it asks. Reward: dev reward.}
\label{tab:bace-gap}
\end{table}

\begin{table}[t]
\centering
\small
\setlength{\tabcolsep}{4pt}
\begin{tabular}{@{}lccccc@{}}
\toprule
\textbf{Draw} & \textbf{Ask rate} & \textbf{Sign agr.} & \textbf{Check} & \textbf{Reward} & \textbf{Under} \\
\midrule
\multicolumn{6}{@{}l}{\textit{Seed 0}} \\
a & .449 & .847 & pass & .905 & .023 \\
b & .390 & .856 & pass & .712 & .531 \\
c & .415 & .883 & pass & .932 & .000 \\
d & .458 & .838 & pass & .881 & .009 \\
\midrule
\multicolumn{6}{@{}l}{\textit{Seed 7}} \\
a & .447 & .835 & pass & .932 & .000 \\
b & .468 & .813 & pass & .898 & .009 \\
c & .947 & .319 & fail & .932 & .000 \\
d & .479 & .802 & pass & .931 & .000 \\
\bottomrule
\end{tabular}
\caption{Four draws of the 8-iteration recipe on each of two seeds. Ask rate and sign agreement of the warm-started chooser, and dev reward and under-asking at the end of training.}
\label{tab:bace-gate}
\end{table}

\section{Belief versus Survivor Set}
\label{app:belief}

$B$ is updated from the transcript with the environment's own rules:
\begin{itemize}
  \item A question about a listed attribute keeps the candidates whose value matches the reply. For available sizes, the reply lists the referent's sizes and $B$ keeps the candidates whose list is the same set.
  \item A question about the half of the scene keeps the candidates on the replied side of a median split on position. A question about a neighbor keeps the candidates that agree with the reply on having a neighbor in that direction. Both use the positions and relations in the observation.
  \item A question the environment can't interpret, a repeated question and a refused one leave $B$ unchanged. A question whose attribute can't be resolved adds nothing to the set $A$ of Equation~\ref{eq:bace-phi}, and a repeated attribute counts once.
\end{itemize}

Besides $\varphi$, each finalist gets three features of its own (Eq.~\ref{eq:bace-psi}). A question gets the entropy its attribute is expected to leave in $B$, whether the attribute is a visual one (always 0 here, since the environment withholds them), and whether it was already asked. The expected entropy uses $\mathrm{IG}_B(a)$, the information gain expected from $a$'s reply when every candidate in $B$ is equally likely, computed with $B$'s own update rules, so a spatial question is credited with the split it would make. An answer gets whether the named object is still in $B$, the chance $1/b$ that a uniform guess from $B$ is right, and a 0, so that both finalists give the Q head the same input width.

\begin{equation}
\begin{aligned}
  \psi_{\text{ask}}(a) &= \big[\, h(a),\ \mathbb{1}[a \text{ visual}],\ \mathbb{1}[a \in A] \,\big],\\
  h(a) &= \min\!\Big(\tfrac{\max(\log_2 b - \mathrm{IG}_B(a),\,0)}{\log_2 \max(n,2)},\,1\Big)\\
  \psi_{\text{ans}}(c) &= \big[\, \mathbb{1}[c \in B],\ \tfrac{1}{b},\ 0 \,\big]
\end{aligned}
\label{eq:bace-psi}
\end{equation}

The feature $h(a)$ of Equation~\ref{eq:bace-psi} uses $\mathrm{IG}_B(a) = \log_2 b - \sum_v \tfrac{b_v}{b}\log_2 b_v$, where $b_v$ counts the candidates in $B$ that $a$'s reply $v$ would keep: those with listed value $v$, those on side $v$ of the median split, or those with or without a neighbor in $B$. Scored on the listed values alone, every spatial question would get zero gain.

To check the belief update of Section~\ref{sec:bace-belief}, we replay every real train, dev and devtest episode (14,525 in all) six times, each time asking three distinct askable attributes in a random order, and compare $B$ with the environment's survivor set before the first question and after each one. That gives 348,600 states per draw, and Table~\ref{tab:belief} reports five draws of the orders. The replay uses no model: each question names its attribute directly, and the reply is the scripted user's.

$B$ differs from the survivor set in 83 to 95 states per draw. All of them come from a handful of episodes: the same 8 in every draw, whose dialogue history already ruled out a candidate (the survivor set starts smaller than the candidate list, while $B$ starts from the full list), and up to 5 more whose candidate list repeats an object id.

During an episode, BACE tracks whether $B$ can still be trusted to equal the survivor set. A reply about a listed attribute that the belief can't parse clears it. Once it is cleared, the updates that need the exact set are skipped for the rest of the episode: the split into halves of the scene, which divides the remaining candidates at the median position, and a ``no'' to a neighbor question, which keeps the candidates without a neighbor in the set. A ``yes'' to a neighbor question and a reply about a listed attribute still narrow $B$. In the replay this happens in 4 to 9 of the 348,600 states.

\begin{table}[t]
\centering
\small
\begin{tabular}{@{}lccc@{}}
\toprule
\textbf{Draw} & \textbf{Not exact} & \textbf{$B \neq S$} & \textbf{Episodes} \\
\midrule
0 & 4 & 83 & 8 \\
1 & 4 & 91 & 12 \\
2 & 8 & 86 & 10 \\
3 & 5 & 95 & 13 \\
4 & 9 & 83 & 9 \\
\bottomrule
\end{tabular}
\caption{Belief against the survivor set $S$ over 348,600 replayed states per draw. Not exact: states where the belief no longer claims to equal $S$. $B \neq S$: states where it does claim to, and differs. Episodes: the episodes those states come from.}
\label{tab:belief}
\end{table}

\section{Question Quality}
\label{app:questions}

Table~\ref{tab:questions} describes the questions each method asks, from the same evaluation runs as Table~\ref{tab:main}. For each question we take the survivors it was asked on and record whether its attribute is spatial (a half of the scene or a neighbor), how many bits of entropy over the survivors it removes, whether it has the highest information gain among the askable attributes, and whether it repeats an attribute already asked in the episode or pinned by the dialogue history.

BACE's questions are close to SFT's, which is what freezing the generator should give: the same mix of attributes, with spatial ones slightly more often, and a best-attribute rate 3 to 5 points lower. The gap most likely comes from how BACE calls the generator: two samples at temperature 0.8 under ArCHer's instruction (Appendix~\ref{app:prompts}), against SFT's greedy decoding under its own. About 3\% of BACE's dev questions repeat an attribute, against 1.5\% for SFT. ArCHer's actor, which RL does train, lands in the same range, and GRPO's questions match SFT's. Natural Behavior's questions are of another kind: 87\% of them rule out nothing, since about half repeat an attribute already asked or pinned by the history and many come on episodes with a single candidate.

\begin{table}[t]
\centering
\small
\setlength{\tabcolsep}{3pt}
\resizebox{\columnwidth}{!}{%
\begin{tabular}{@{}lcccccc@{}}
\toprule
\textbf{Method} & \textbf{Qs asked} & \textbf{Spatial} & \textbf{Bits} & \textbf{Best attr.} & \textbf{Repeat} & \textbf{No split} \\
\midrule
\multicolumn{7}{@{}l}{\textit{Dev}} \\
Natural Behavior & 758 & .029 & 0.18 & .127 & .496 & .865 \\
SFT & 460 & .265 & 1.65 & .858 & .015 & .007 \\
GRPO & 443 & .234 & 1.68 & .843 & .024 & .001 \\
ArCHer & 452 & .237 & 1.62 & .799 & .016 & .032 \\
BACE & 473 & .309 & 1.57 & .815 & .032 & .000 \\
\midrule
\multicolumn{7}{@{}l}{\textit{Devtest}} \\
Natural Behavior & 2,267 & .056 & 0.13 & .131 & .483 & .875 \\
SFT & 1,059 & .304 & 1.54 & .881 & .013 & .015 \\
GRPO & 1,008 & .250 & 1.59 & .867 & .028 & .002 \\
ArCHer & 1,002 & .258 & 1.57 & .856 & .021 & .021 \\
BACE & 1,072 & .324 & 1.50 & .846 & .037 & .005 \\
\bottomrule
\end{tabular}
}
\caption{Questions per method, mean over seeds (BACE: seeds 0, 1 and 7). Qs asked: questions on the split. Spatial: share about a half of the scene or a neighbor. Bits: mean entropy removed from the survivor set. Best attr.: share whose attribute has the highest information gain. Repeat: share that repeats an attribute already asked or pinned by the history. No split: share that rules out no survivor.}
\label{tab:questions}
\end{table}

\section{Example Episodes}
\label{app:examples}

\begin{figure}[!t]
  \centering
  \includegraphics[width=\columnwidth]{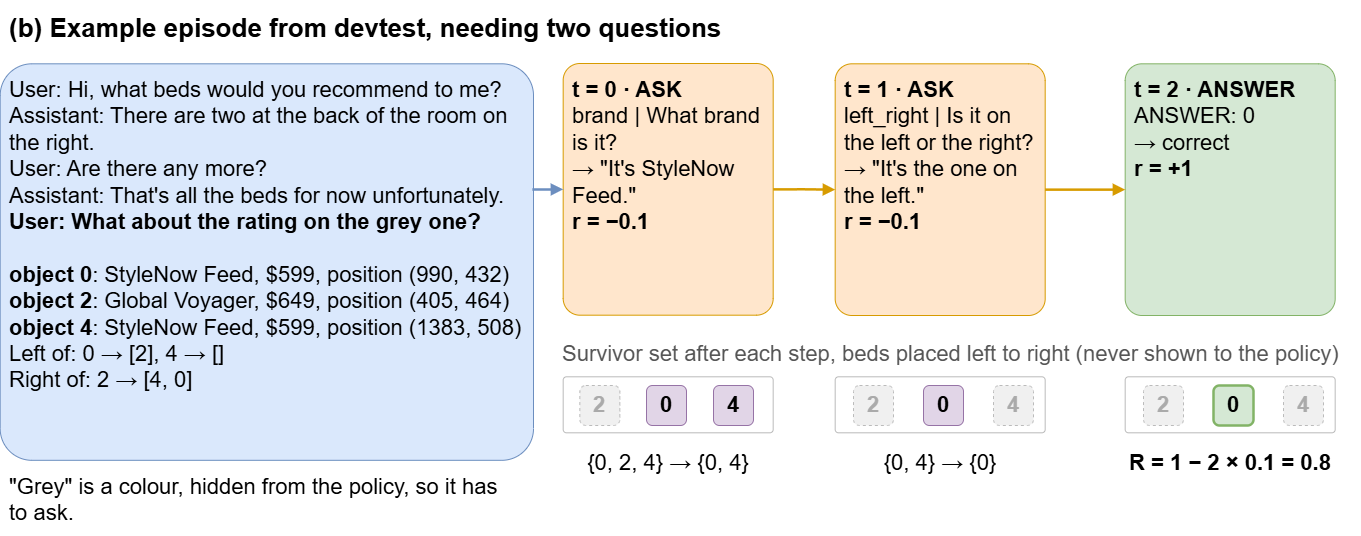}
  \caption{A devtest episode that needs two questions: ``the grey one'' names a color the policy can't see, so it asks for the brand and then for the side of the scene before answering.}
  \label{fig:task-example}
\end{figure}

Figure~\ref{fig:task-example} follows one devtest episode through the task of Figure~\ref{fig:task}. The evaluation records keep each policy's attribute and the survivors after every reply, but not the wording of its questions, so the traces below give the attribute asked and how many candidates were left after the reply. All are dev episodes, with seed 7 for every trained method, each at the end of its pass.

In the first episode the user asks ``Can you give me the sizes for the gray jacket on the front rack and the and for the red and white jacket?'' with eight candidates, five of which are the same Coats \& More jacket (sizes XS and L, rated 3.3, \$234.99). The referent is object 11, and the oracle needs three questions.
\begin{itemize}
  \item SFT and BACE: brand (8 left $\to$ 5), half of the scene left/right ($\to$ 2), half of the scene top/bottom ($\to$ 1), then answer object 11, correct.
  \item GRPO and ArCHer: brand ($\to$ 5), left/right ($\to$ 2), then answer object 34 with two candidates still possible, wrong.
\end{itemize}
This is the most common error of GRPO and ArCHer on ask-2 and ask-3 episodes (Section~\ref{sec:res-main}).

The second shows BACE failing. After a dialogue about hoodies in size M, the user asks ``How much does that one over there cost?'' with 39 candidates; the oracle needs two questions. SFT asks about customer reviews ($\to$ 2) and available sizes ($\to$ 1) and answers correctly. BACE asks about the left/right half ($\to$ 20), the top/bottom half ($\to$ 10) and the left/right half again, which the environment treats as a repeat and answers without narrowing, and then answers with ten candidates left, wrongly. Asking a third time was the right call with ten candidates left; what went wrong is that the sampled questions split the candidates weakly and the last one repeated an attribute. In another episode (``Let's go back to coats. What is the review and brand of that one?'', nine candidates), none of BACE's three sampled questions maps to an attribute, so nothing is ruled out and its answer is wrong; SFT resolves the same episode in three questions.

On an unambiguous episode, Natural Behavior replies ``Great! I've added the black coat in size L to your cart. Is there anything else you need help with?'' The reply ends with a question mark and contains ``size'', so it counts as a question about size on an episode that needs none: the over-asking of Table~\ref{tab:main-dev}. On another unambiguous episode it replies ``Sure, I'll add the grey jeans to your cart. Is there anything else you need help with?'' This also counts as a question, but it matches no attribute, so the scripted user answers ``Sorry, I didn't understand the question. Could you rephrase it?'' and the turn costs $\lambda$ without ruling anything out. Counting phrases such as ``anything else'', ``to your cart'' or ``would you like me to'', 234 of the 368 first-turn replies counted as questions on dev are offers or follow-ups of this kind, and 6 ask the user which object they mean.

\section{Dataset Construction Details}
\label{app:dataset}

Table~\ref{tab:conditions} shows how the labelled user turns become episodes. A turn whose mention SIMMC~2.1 doesn't mark as ambiguous has one annotated referent and becomes an unambiguous episode. An ambiguous mention whose dialogue history leaves exactly one candidate is recoverable and also needs no question; this happens five times in train and never in dev or devtest. An ambiguous mention that leaves two or more candidates needs a question. We drop ambiguous mentions with fewer than two annotated candidates, turns whose history leaves no candidate, and a few with inconsistent labels, such as a gold referent outside its own survivor set.

The history constraints are the attribute values SIMMC~2.1 annotates on each turn, such as a brand the user asked for. They are carried forward only while the dialogue stays on the same reference: every turn is tagged with the objects it is about (its candidate list, or the object it resolves to), and when that set changes we clear the constraints. Without this reset, an attribute the user mentioned for a different item would filter the candidates of a later, unrelated mention in the same scene.

The observation gives each candidate's position, the centre of its bounding box, and the scene's relation graph, which lists the objects to the left of, right of, above and below each one. A question about the half of the scene ranks the candidates still in play by position along that axis and splits them at the median, and the reply says which half holds the referent. A question about a neighbor (``is there another one to its left?'') asks whether the referent has a neighbor in that direction among the candidates still in play, and the reply keeps the candidates that agree. Both are computed on the current survivors, so the same question can split a different set later in the episode.

Table~\ref{tab:conditions} also gives, for the ambiguous episodes of each split, how the gold referent was recovered: from a later turn that resolves the mention to a single candidate, from one that narrows it to two (we take the first), or, when no later turn narrows the set, by a seeded random choice among the candidates. The shares are similar across splits. Appendix~\ref{app:single-gold} scores the policies on the single-gold and fallback episodes separately.

An episode is unresolvable when the askable attributes, each asked once, run out before one candidate is left, for example when several candidates share every listed value and the spatial questions can't tell them apart. There are 47 in train, 1 in dev and 12 in devtest. No policy can be sure to win them, so we leave them out of training and keep them in evaluation, in the ask-3 group. Another 15, 3 and 3 episodes need four or five questions, more than the budget, and are also in ask-3.

A question past the budget of three is refused, and the policy must answer; if it asks again, the episode ends and the environment returns a random candidate from the survivor set still consistent with the replies so far.

\begin{table}[t]
\centering
\small
\setlength{\tabcolsep}{3pt}
\resizebox{\columnwidth}{!}{%
\begin{tabular}{@{}lrrr@{}}
\toprule
 & \textbf{Train} & \textbf{Dev} & \textbf{Devtest} \\
\midrule
Labelled turns & 11,753 & 1,050 & 2,657 \\
Dropped & 671 & 90 & 174 \\
Unambiguous & 7,400 & 617 & 1,701 \\
Recoverable from history & 5 & 0 & 0 \\
Needs a question & 3,677 & 343 & 782 \\
\midrule
\multicolumn{4}{@{}l}{\textit{Gold referent of the ambiguous episodes}} \\
Later turn, one object & 1,779 (48\%) & 150 (44\%) & 380 (49\%) \\
Later turn, two objects & 1,070 (29\%) & 106 (31\%) & 248 (32\%) \\
Seeded random choice & 828 (23\%) & 87 (25\%) & 154 (20\%) \\
\bottomrule
\end{tabular}
}
\caption{From labelled user turns to episodes, and how the gold referent of each ambiguous episode was recovered.}
\label{tab:conditions}
\end{table}

\section{Episodes with a Single Gold Referent}
\label{app:single-gold}

The gold referent of an ambiguous episode is a single object named in a later turn for 49\% of the devtest episodes and 44\% of the dev ones (Table~\ref{tab:conditions}). For the others it is one of two objects a later turn narrows the mention to, or a seeded random choice among the candidates. The scripted user answers every question about the gold object, so these episodes can still be solved, but that object may not be the one the user meant. Table~\ref{tab:single-gold} scores the policies of Table~\ref{tab:main}, at the same points, on each kind of ambiguous episode separately. The numbers come from the same evaluation runs restricted to these episodes. Ask F1 and over-asking are left out, since neither set contains unambiguous episodes.

\begin{table}[t]
\centering
\small
\setlength{\tabcolsep}{3pt}
\resizebox{\columnwidth}{!}{%
\begin{tabular}{@{}lcccccc@{}}
\toprule
 & \multicolumn{3}{c}{\textbf{Single gold}} & \multicolumn{3}{c}{\textbf{Fallback gold}} \\
\cmidrule(lr){2-4} \cmidrule(lr){5-7}
\textbf{Method} & \textbf{Hit} & \textbf{Acc.} & \textbf{Reward} & \textbf{Hit} & \textbf{Acc.} & \textbf{Reward} \\
\midrule
\multicolumn{7}{@{}l}{\textit{Devtest, 380 and 402 episodes}} \\
Natural Behavior & .176 & .329 & $-$.402 & .075 & .289 & $-$.451 \\
SFT (start of RL) & .953 & .976 & .825 & .925 & .963 & .783 \\
GRPO & .907\sd{.060} & .961\sd{.033} & .802\sd{.058} & .863\sd{.108} & .930\sd{.057} & .730\sd{.101} \\
ArCHer & .901\sd{.026} & .956\sd{.012} & .788\sd{.022} & .833\sd{.057} & .913\sd{.028} & .694\sd{.051} \\
BACE & .953\sd{.000} & .976\sd{.000} & .824\sd{.000} & .928\sd{.000} & .971\sd{.001} & .797\sd{.003} \\
\midrule
\multicolumn{7}{@{}l}{\textit{Dev, 150 and 193 episodes}} \\
Natural Behavior & .127 & .400 & $-$.255 & .098 & .301 & $-$.427 \\
SFT (start of RL) & .947 & .973 & .819 & .938 & .953 & .766 \\
GRPO & .745\sd{.417} & .817\sd{.304} & .537\sd{.554} & .719\sd{.402} & .806\sd{.309} & .505\sd{.558} \\
ArCHer & .889\sd{.035} & .960\sd{.018} & .796\sd{.033} & .859\sd{.053} & .923\sd{.019} & .709\sd{.032} \\
BACE & .967\sd{.000} & .977\sd{.004} & .827\sd{.006} & .927\sd{.000} & .969\sd{.000} & .793\sd{.000} \\
\bottomrule
\end{tabular}
}
\caption{The ambiguous episodes split by how the gold referent was recovered (Table~\ref{tab:conditions}). Single gold: a later turn names one object. Fallback gold: a later turn narrows the mention to two objects, or a seeded random choice. Points and metrics as in Table~\ref{tab:main}. Mean over seeds, SD in parentheses.}
\label{tab:single-gold}
\end{table}

\end{document}